\documentclass{article}

\usepackage[nonatbib,final]{neurips_2023}

\usepackage[utf8]{inputenc} 
\usepackage[T1]{fontenc}    
\usepackage[hidelinks]{hyperref}       
\usepackage{url}            
\usepackage{booktabs}       
\usepackage{amsfonts}       
\usepackage{nicefrac}       
\usepackage{microtype}      
\usepackage{xcolor}         
\usepackage{amsmath}
\usepackage{soul}
\usepackage{graphicx}
\usepackage{subcaption}
\usepackage{cinzel} 
\usepackage{multirow}
\usepackage{multicol}
\usepackage{tabularx}

\usepackage{comment} 
\usepackage{wrapfig} 
\usepackage{bbm}
\usepackage{pifont}
\usepackage{makecell,rotating}
    \setcellgapes{5pt}

\usepackage{calligra}

\usepackage{aurical}

\usepackage{listings}
\usepackage{color}
\usepackage[T1]{fontenc}
\usepackage{numprint}

\npthousandsep{\,}
\npdecimalsign{.}
\npproductsign{\cdot}
\npunitseparator{}

\definecolor{dkgreen}{rgb}{0,0.6,0}
\definecolor{gray}{rgb}{0.5,0.5,0.5}
\definecolor{mauve}{rgb}{0.58,0,0.82}

\usepackage{mathtools}
\usepackage{titlesec}
\titlespacing*{\section}
{0pt}{0ex}{0ex}
\titlespacing*{\subsection}
{0pt}{0ex}{0ex}
\titlespacing*{\subsubsection}
{0pt}{0ex}{0ex}
\usepackage[export]{adjustbox}

\newenvironment{tight_enumerate}{
\begin{enumerate}[leftmargin=*]
  \setlength{\itemsep}{0pt}
  \setlength{\parskip}{0pt}
}{\end{enumerate}}
\newenvironment{tight_itemize}{
\begin{itemize}
  \setlength{\itemsep}{0pt}
  \setlength{\parskip}{0pt}
}{\end{itemize}}

\newenvironment{tightcenter}{%
  \setlength\topsep{0pt}
  \setlength\parskip{0pt}
  \begin{center}
}{%
  \end{center}
}

\makeatletter
\def\thickhline{%
  \noalign{\ifnum0=`}\fi\hrule \@height \thickarrayrulewidth \futurelet
   \reserved@a\@xthickhline}
\def\@xthickhline{\ifx\reserved@a\thickhline
               \vskip\doublerulesep
               \vskip-\thickarrayrulewidth
             \fi
      \ifnum0=`{\fi}}
\makeatother

\makeatletter
\newcommand\footnoteref[1]{\protected@xdef\@thefnmark{\ref{#1}}\@footnotemark}
\makeatother

\newlength{\thickarrayrulewidth}
\usepackage{titlesec}

\titleformat{\paragraph}
{\normalfont\normalsize\bfseries}{\theparagraph}{1em}{}
\titlespacing*{\paragraph}
{0pt}{3.25ex plus 1ex minus .2ex}{1.5ex plus .2ex}

\newcommand{\DyadDim}{${n}_{dyad}$}
\newcommand{\DimIn}{${n}_{in}$}
\newcommand{\DimOut}{${n}_{out}$}

\newcommand{\LayerGroupName}{Dyad}
\newcommand{\LayerVariantNameI}{Dyad-IT}
\newcommand{\LayerVariantNameII}{Dyad-OT}
\newcommand{\LayerVariantNameIII}{Dyad-DT}

\newcommand{\DyadComponentI}{Block Diagonal Component}
\newcommand{\DyadComponentII}{Block Transposed Component}
\newcommand{\DyadComponentIShorthand}{\textsc{BlockDiag}}
\newcommand{\DyadComponentIIShorthand}{\textsc{BlockTrans}}

\newcommand{\titlecolor}{violet}
\title{\textcinzelblack{PINEAPPLE}: {\textcolor{\titlecolor}P}ersonifying {\textcolor{\titlecolor} I}{\textcolor{\titlecolor} N}animate {\textcolor{\titlecolor}E}ntities by {\textcolor{\titlecolor}A}cquiring {\textcolor{\titlecolor}P}arallel {\textcolor{\titlecolor}P}ersonification Data for {\textcolor{\titlecolor}L}earning {\textcolor{\titlecolor}E}nhanced Generation}

\title{DYAD: A Descriptive Yet Abjuring Density efficient approximation to linear neural network layers}

\author{%
Sarin Chandy\thanks{Equal Contribution. Sarin proposed the dyad model, modified the models in the transformers library with the Dyad layer and wrote the formulation section. Varun proposed the evaluation frameworks, organized the experiments and led the paper writing} \quad Varun Gangal \footnotemark[1] \quad Yi Yang \quad Gabriel Maggiotti  \\
ASAPP Inc.\\
\texttt{\{schandy,vgangal,yyang,gmaggiotti\}@asapp.com}\\
}

\begin{document}

\maketitle

\begin{abstract}
We devise, implement and performance-asses \textsc{\LayerGroupName}, a layer which can serve as a faster and more memory-efficient approximate replacement for linear layers, (\textit{nn.Linear()} in Pytorch). These layers appear in common subcomponents, such as in the \textit{ff} module of Transformers. \textsc{\LayerGroupName} is based on a bespoke near-sparse matrix structure which approximates the dense "weight" matrix $W$ that matrix-multiplies the input in the typical realization of such a layer, a.k.a \textsc{Dense}. Our alternative near-sparse matrix structure is decomposable to a sum of 2 matrices permutable to a block-sparse counterpart. These can be represented as 3D tensors, which in unison allow a faster execution of matrix multiplication with the mini-batched input matrix $X$ compared to \textsc{Dense} $(O(rows(W) \times cols(W)) \rightarrow O(\frac{rows(W) \times cols(W)}{\#{\ }of{\ }blocks}))$. As the crux of our experiments, we pretrain both \textsc{\LayerGroupName} and \textsc{Dense} variants of 2 sizes of the OPT arch and 1 size of the Pythia arch, including at different token scales of the babyLM benchmark. We find \textsc{\LayerGroupName} to be competitive ($\geq$ $90$\%) of \textsc{Dense} performance on zero-shot (e.g. \textsc{Blimp}), few-shot (\textsc{OpenLM}) and finetuning (\textsc{GLUE}) benchmarks, while being $\geq$7-15\% faster to train on-GPU even at 125m scale, besides surfacing larger speedups at increasing scale and model width.

\end{abstract}
\vspace*{-1.02\baselineskip}

\section{Introduction}
\label{sec:intro}
Riding on the back of  the already pivotal decade-long rise of GPU-driven deep learning \cite{krizhevsky2012imagenet},  Transformers \cite{vaswani2017attention} in 2017 crescendoed the ambition, scale and task-generality of ML models. With cross-sequence in-training parallelizability and representation power through all-pair interactions, transformers disrupted NLP and its incumbent recurrent paradigm \cite{sutskever2014sequence}, but since became key components in other modalities such as CV \cite{srinivas2021bottleneck}. Pretrained models as base representations, limited then to CV, emerged via LLMs like BERT \cite{devlin2018bert}, T5 \cite{raffel2020exploring} etc reaching SOTA across tasks with limited finetuning. 

A natural consequence of a module's ubiquity is that even a small improvement to one of its aspect can have major impact on its application and research --- as seen by the recent impact of e.g., quantization \cite{dettmers2023qlora}. A result of this is that an inefficient component (attention) sees a barrage of research (e.g.hashing \cite{kitaev2020reformer}, softmax alternatives \cite{qin2022cosformer}, FlashAttention \cite{dao2022flashattention} etc) until some other component emerges as a bottleneck. We believe this is the case with the dense linear layers in the Transformer's $ff$ module. Moreover, models have larger hidden dimension (4096 for Pythia, 8192 for Llama2), leading to quadratic rise in compute from \textit{ff} module linear layers. Thus inspired, we devise DYAD (\textit{Descriptive Yet Abjuring Density}) --- an efficient linear layer approximation using block-sparsity.

We also release the better fraction of our code at \url{https://github.com/asappresearch/dyad}

\section{Formulation}
\label{sec:formulation}
\subsection{Linear Layer}
A Linear layer is the basic building block of all neural networks, represented in pytorch by \verb|nn.Linear()|. It maps input $X$ to output $Y$ via a dense matrix multiplication with weight matrix $W$, given by the equation $Y = G_{Linear}(X) = WX+b$. Here, $W$ is a matrix of shape $f_{out} \times f_{in}$ where $f_{out}$ and $f_{in}$ represent the no. of output \& input features. $Y$, $X$ and the bias $b$ have shapes $f_{out} \times n_{batch}$, ${f}_{in} \times n_{batch}$ and $f_{out} \times 1$. Frameworks like pytorch pose the shape of $X$ and $Y$ as $n_{batch} \times f_{in}$ and $n_{batch} \times f_{out}$ but here we adhere to the former convention.

\subsection{\textsc{\LayerGroupName} : Definition and Properties}
We introduce a family of sparse layers named \textsc{\LayerGroupName} that can serve as an approximate replacement for the dense linear layer. \textsc{\LayerGroupName} has 3 variants called \textsc{\LayerVariantNameI}, \textsc{\LayerVariantNameII} and \textsc{\LayerVariantNameIII}. The initials stand for Input Transpose, Output Transpose and Double Transpose. They are named such because transpose operations on either the input or output enables to compute their outputs efficiently. We describe \textsc{\LayerVariantNameI} here and will describe the other two in a later section.
\begin{figure*}[t]
    \centering
    \includegraphics[width=0.25\textwidth,frame]{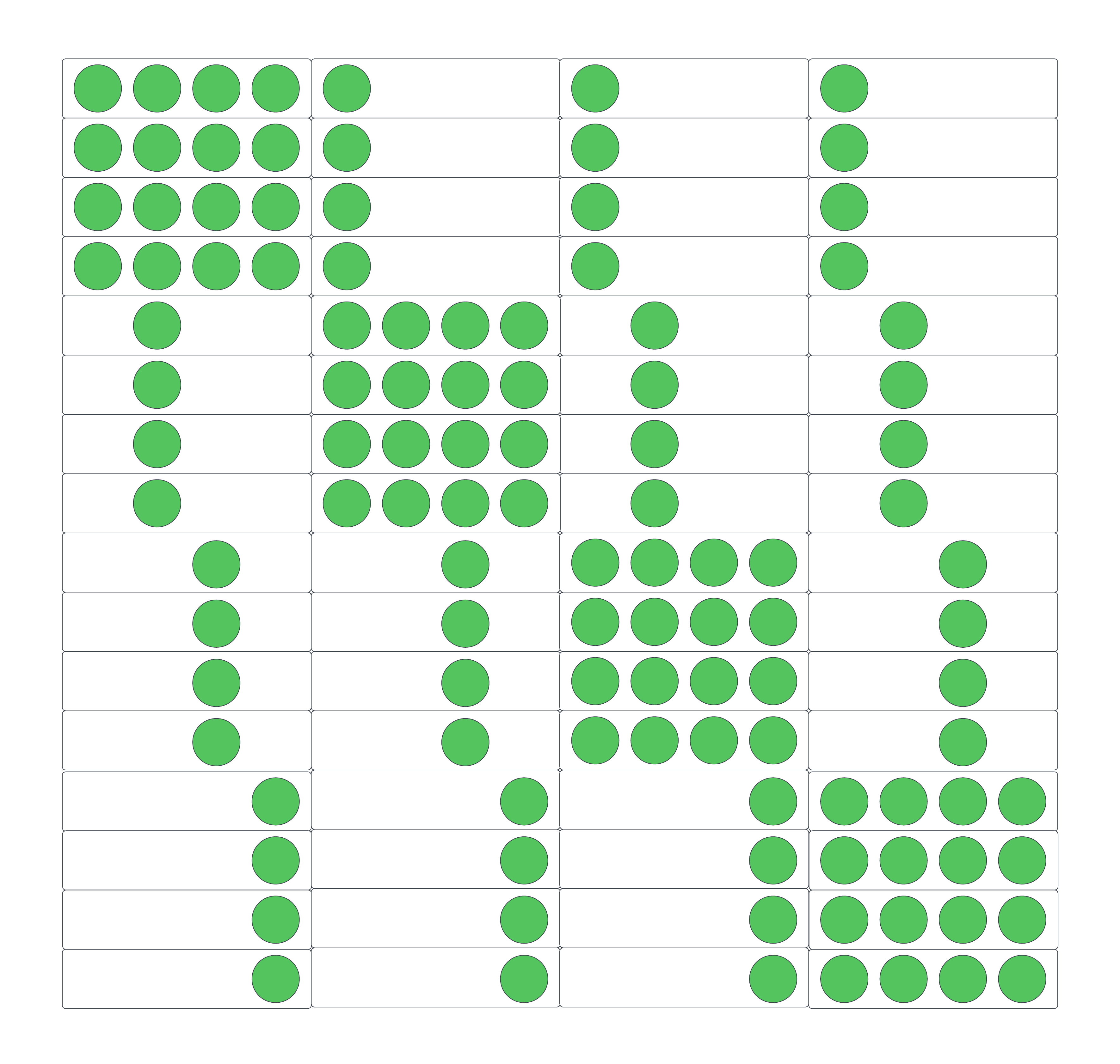}
    \includegraphics[width=0.52\textwidth,frame]{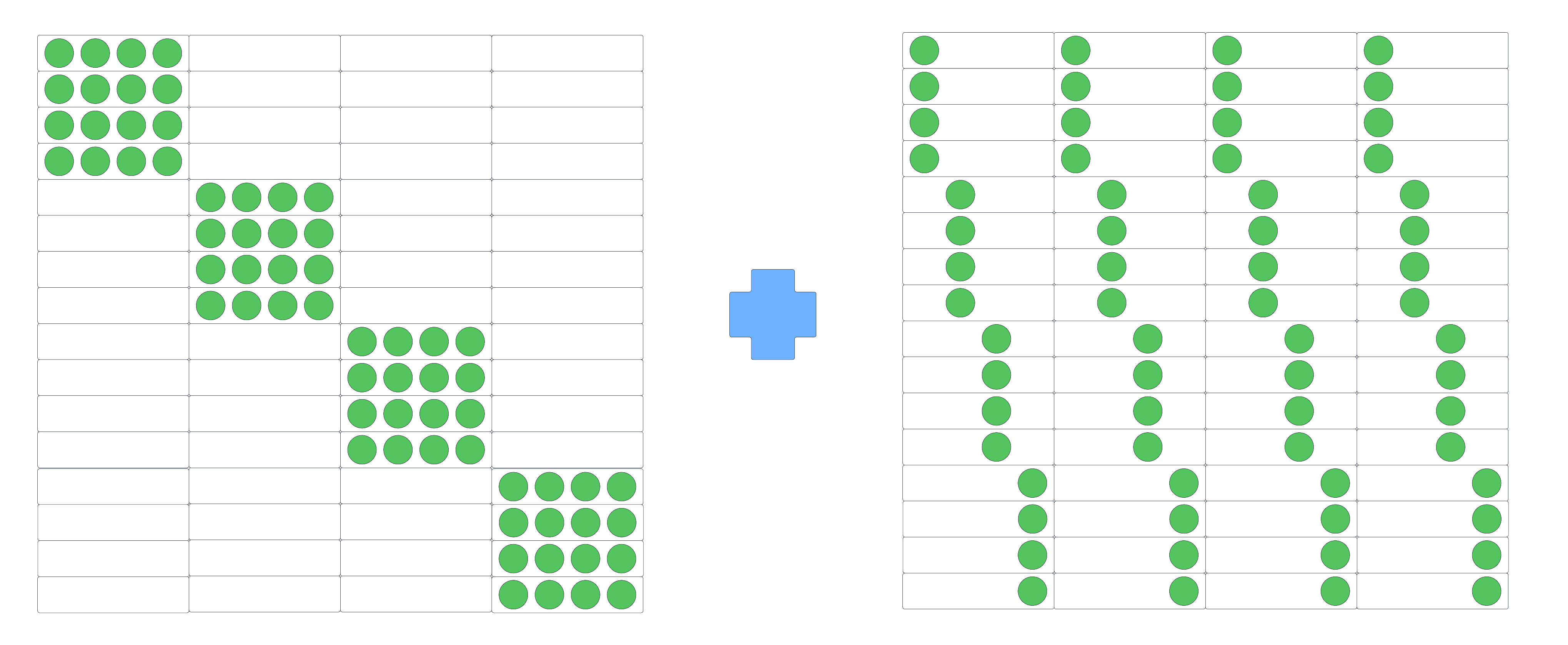}
    \caption{\footnotesize Dyad Weight Matrix [L] vs its Components [R], {\DyadComponentIShorthand} \& {\DyadComponentIIShorthand}. Green is $\neq 0$. } \label{fig:dyad-input-transpose-components}
\vspace{-3mm}
\end{figure*}
\textsc{\LayerGroupName} is a linear layer with a sparse weight matrix having shape shown in Fig \ref{fig:dyad-input-transpose-components}. The output of this layer can be calculated using $G_{Linear}$. However, this won't lead to any efficiency gain compared to the linear layer. We can split the \textsc{\LayerGroupName} matrix into 2 components as shown in Fig \ref{fig:dyad-input-transpose-components}. These components share some non-zero elements but their sum's representational power would be identical to the \textsc{\LayerGroupName} matrix. We call the first component the \textit{\DyadComponentI} ({\DyadComponentIShorthand}) and the second one the \textit{\DyadComponentII} ({\DyadComponentIIShorthand}). The ability to split \textsc{\LayerGroupName} into 2 components is what inspires its name. A \textsc{\LayerGroupName} matrix can be defined using 3 parameters, \textsc{\DyadDim}, \textsc{\DimIn} and \textsc{\DimOut}. \textsc{\DimOut} $\times$ \textsc{\DimIn} is the size of each submatrix in \textit{\DyadComponentIShorthand} and \textsc{\DyadDim} represents the no. of submatrixes in each component. Thus, all the figures for \textsc{\LayerGroupName} shown here have $n_{dyad}=n_{in}=n_{out}=4$. With the 2 components of \textsc{\LayerGroupName} split up, we can write its layer output as in Eq \ref{dyadFullEq}.
\begin{align}
    Y &= W_1X + W_2X + b \label{dyadFullEq}
\end{align}
Naively implementing this as in Eq \ref{dyadFullEq}, will be as expensive as its dense counterpart. To exploit the joint properties of sparsity and block structure in these 2 components, we need to transform $W_1X$ and  $W_2X$ to an equivalent sequence of 3D tensor operands and operations.

Hereforth, we ease representing 3D tensors in our equations by overloading pytorch tensor operators.

\subsubsection{Efficient Computation of {\DyadComponentIShorthand}}
Let $Y_{1} = W_1X$ be the output of {\DyadComponentIShorthand}. From Fig \ref{fig:dyad-input-transpose-components}, we can see that for any $Y_{1}[i\times{{n}_{out}}:(i+1)\times{{n}_{out}},:]$ only depends on $X[i\times{{n}_{in}}:(i+1)\times{{n}_{in}},:]$ where $i \in$ $[0,$\textsc{\DyadDim}$)$. This shows that each pair of $Y_{1}[i \times {{n}_{out}}: (i + 1) \times {{n}_{out}},:]$,$X_{1}[i \times {{n}_{in}}: (i + 1) \times {{n}_{in}},:]$ can be calculated individually using a matrix multiplication. The weights needed for this are $W_{1}[i\times{{n}_{out}}:(i+1)\times{{n}_{out}},i\times{{n}_{in}}:(i+1)\times{{n}_{in}}]$. We can store the weights needed for all these pairs of outputs and inputs as a 3D tensor,$W_1^{'}$ of shape (\textsc{\DyadDim},\textsc{\DimOut},\textsc{\DimIn}) as per Eq \ref{DyadMap}.
{\small\begin{align}
    W_1^{'}[i,j,k] &= W_1[i*{{n}_{out}}+j,i*{{n}_{in}}+k] \label{DyadMap}
\end{align}}%
This is a factor of \textsc{\DyadDim} times smaller when compared to $W_1$ since it has the shape (\textsc{\DyadDim}$\times$\textsc{\DimOut} , \textsc{\DyadDim}$\times$\textsc{\DimIn}). Thus, the whole output of the layer can be computed together with a single batched matrix multiplication as shown in Eq \ref{diagBMM} after the input has been also converted to a 3D tensor as shown in Eq \ref{diagReshape}.
{\small\begin{align}
    X_1^{'} &= X.reshape({{n}_{dyad}},{{n}_{in}},n_{batch}) \label{diagReshape}\\
    Y_1 &= W_1^{'}.bmm(X_1^{'}).reshape({{n}_{dyad}} \times {{n}_{out}},{n}_{batch}) \label{diagBMM}
\end{align}}%
The value of $Y_1$ here is the same as $W_1X$ but cost of computing it will be $O({{n}_{dyad}} \times {{n}_{out}} \times {{n}_{in}})$ instead of $O({n}_{dyad}^{2} \times {{n}_{out}} \times {{n}_{in}})$ which is $O({{n}_{dyad}})$ times faster.

\subsubsection{Efficient Computation of {\DyadComponentIIShorthand}}
The matrix multiplication for {\DyadComponentIIShorthand}, i.e. $W_2X$ can be converted to a form similar to {\DyadComponentIShorthand} by permuting the columns of $W_2$. A permutation matrix, $P$ is a square matrix which has exactly one element along each row and each column as one and the rest have a value of zero. Pre-multiplying by a permutation matrix ($PA$), permutes the rows of matrix $A$, while post-multiplying ($AP$), permutes the columns of matrix $A$. So if we post multiply, $W_2$ by an appropriate permutation matrix which has the form as shown in Eq \ref{PermuteEq}, we will end up with a matrix similar to {\DyadComponentIShorthand}. 
\begin{align}
    P(i,j) &= \delta_{j={{n}_{dyad}}*(i\%{{n}_{in}}) + i//{{n}_{in}}} \label{PermuteEq} 
\end{align}
\begin{wrapfigure}{l}{0.34\textwidth}
    \includegraphics[width=0.99\linewidth]{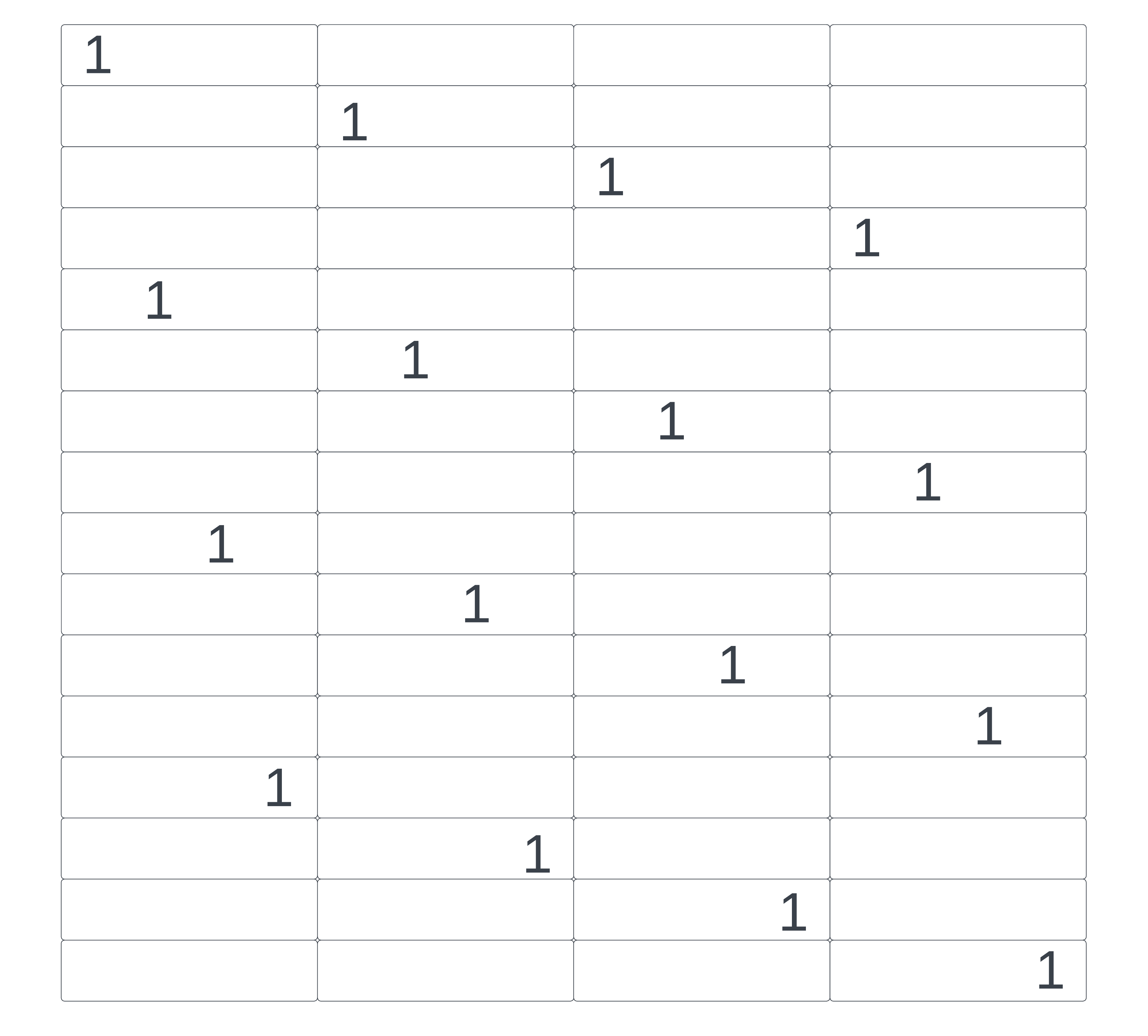}
    \caption{Permutation Matrix for \DyadComponentIIShorthand}
    \label{fig:permutation-matrix}
\end{wrapfigure}
For {\LayerGroupName} with $n_{dyad}=n_{in}=n_{out}=4$, this permutation matrix is shown in Fig \ref{fig:permutation-matrix}. As Permutation matrices are orthonormal, $P^{-1} = P^{T}$ where $P^{T}$ is another Permutation matrix \cite{wiki-matrix-perm-properties}. 
{\footnotesize\begin{align}
    Y_2 = W_2X = W_2(PP^{T})X = (W_2P)(P^{T}X) \label{permuteComp}
\end{align}}%
Using the othonormal property of permutation matrixes we can write $W_2X$ as shown in Eq \ref{permuteComp}. Here, $Y_2$ is the output of {\DyadComponentIIShorthand}. Let $W_2^{P} = W_2P$ and $X_2^{P} = P^{T}X$. Since, $W_2^{P}$ has the same structure as the weight matrix for {\DyadComponentIShorthand}, we can also store it as a 3D tensor,$W_2^{'}$, of shape ${{n}_{dyad}} \times {{n}_{out}} \times {{n}_{in}}$. Hence, as in the case before this leads to a reduction in memory size of $O({{n}_{dyad}})$ when compared to $W_2$.


Calculating $P^{T}X$ naively in order to get $X_2^{P}$ will be as expensive as the linear layer. However, the specific pattern of the permutation allows us to calculate $X_2^{P}$ by simplying transposing a 3D view of $X$. We can see from Eq \ref{PermuteEq} that within a multiple of {\DimIn} for every increment of $i$, $j$ increases by {\DyadDim}, while for every $i + {{n}_{in}}$ increment j only increases by 1. Thus, $i$ can be thought of as the 1D index of a flattened 2D matrix of shape ${{n}_{dyad}} \times {{n}_{in}}$ with stride $(1, {{n}_{dyad}})$. Hence, permuting and inverting the permutation can be done by just transposing this 2D matrix i.e. going from shape ${{n}_{dyad}} \times {{n}_{in}}$ to ${{n}_{in}} \times {{n}_{dyad}}$ and from stride $(1, {{n}_{dyad}})$ to stride $({{n}_{dyad}},1)$ and vice versa. Calculating $X_2^{P}$ this way from $X$ is shown in Eq \ref{Xpermute}.

{\footnotesize\begin{align}
    X_2^{P} &= X.reshape({{n}_{in}},{{n}_{dyad}},-1).transpose(0,1).reshape(-1, {n}_{batch}) \label{Xpermute} \\
    X_2^{'} &= X_2^{P}.reshape({{n}_{dyad}},{{n}_{in}},{n}_{batch}) \label{X3d} \\
    X_2^{'} &= X.reshape({{n}_{in}},{{n}_{dyad}},{n}_{batch}).transpose(0,1) \label{Xcomp2Eq}
\end{align}}%

\begin{figure*}[t]
    \centering
    \includegraphics[width=0.31\textwidth]{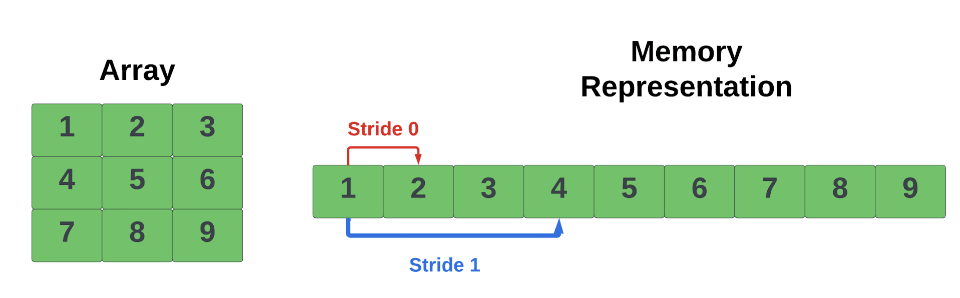}
    \includegraphics[width=0.31\textwidth]{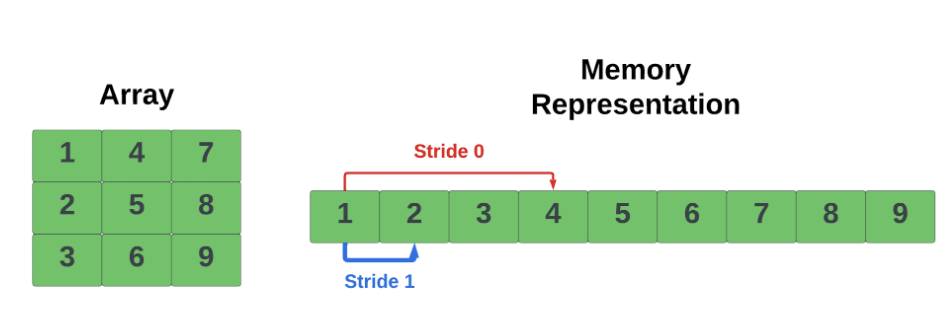}
    \caption{\scriptsize Illustrations of {\DyadComponentIIShorthand} computation, in particular the Equation \ref{Xcomp2Eq} step } \label{fig:blocktranscomputation}
\vspace{-3mm}
\end{figure*}

Now as in the case of {\DyadComponentIShorthand} we need the activations as a 3D tensor to do the calculations efficiently. So we need to reshape $X_2^{P}$ as shown in Eq \ref{X3d} to get $X_2^{'}$. $X_2^{'}$ is the actual activation input for the batched matrix multiplication. We can combine Eq \ref{Xpermute} and \ref{X3d} to cancel the reshape as shown in \ref{Xcomp2Eq}. Eq \ref{Xcomp2Eq} is basically free and involves just changing some metadata related to the strides of the dimensions, as shown in Fiture \ref{fig:blocktranscomputation}. The actual data of the tensor need not be touched here. 
\begin{align}
    Y_2 &= W_2^{'}.bmm(X_2^{'}).reshape({{n}_{dyad}} \times {{n}_{out}}, {n}_{batch}) \label{comp2eq}
\end{align}
Finally, the output of {\DyadComponentIIShorthand} can be computed as shown in Eq \ref{comp2eq}. The cost of computation here is thus, the same as that of the previous case with complexity $O({{n}_{dyad}} \times {{n}_{out}} \times {{n}_{in}})$ and is faster by a factor of $O({{n}_{dyad}})$ when compared with multiplying with $W_2$. In Appendix \S 5.3.3, we discuss some thoughts about the representational power of \textsc{\LayerGroupName}.

\subsection{\textsc{\LayerGroupName} implementation in pytorch}
\label{subsec:PytorchImplementation}
The \textsc{\LayerGroupName} layer can be written relatively efficiently with a few lines of code in native pytorch. Here we present a simple implementation of \textsc{\LayerGroupName}, more specifically the exemplary \textsc{\LayerVariantNameI}. Note that, in the code, we use $dim$ instead of $n$ to denote dimension. We also note that this code has some overhead in terms of multiple kernel launches, sequential processing of the components and some copying that could be avoided. Even with all of this we are still able to observe significant speedups especially at higher model scales.
\begin{lstlisting}
class Dyad(torch.nn.Module):
    def __init__(self,shape,bias=True):
        super().__init__()
        self.dyad_dim, self.dim_in, self.dim_out = shape
        self.has_bias = bias
        k = 1.0/float(np.sqrt(dim_in*dyad_dim))
        self.wu = torch.nn.Parameter(torch.empty((dyad_dim,dim_out,dim_in)))
        torch.nn.init.uniform_(self.wu,-k,k)
        self.wl = torch.nn.Parameter(torch.empty((dyad_dim,dim_out,dim_in)))
        torch.nn.init.uniform_(self.wl,-k,k)
        if self.has_bias:
            self.bias = torch.nn.Parameter(torch.empty((dyad_dim*dim_out,1)))
            torch.nn.init.uniform_(self.bias,-k,k)

    def forward(self,x):
        # The shape of x is (dyad_dim x dim_in, batch_size)
        x1 = x.reshape(self.dyad_dim,self.dim_in,-1)
        # The shape of x1, which is a view of x, is now (dyad_dim, dim_in, batch_size)
        x2 = x.reshape(self.dim_in,self.dyad_dim,-1).transpose(0,1)
        out = (self.wl.bmm(x1)+self.wu.bmm(x2)).reshape(self.dyad_dim*self.dim_out,-1)
        if self.has_bias:
            out+= self.bias
        return out
\end{lstlisting} \label{code:dyadCode}

\subsection{{\LayerGroupName} Variants}

In this section we will describe the other two variants of {\LayerGroupName}, {\LayerVariantNameII} and {\LayerVariantNameIII}. Both of these variants can be split into two components. As in the case of {\LayerVariantNameI}, the first component is a block diagonal matrix and the second component can be converted back into a block diagonal by means of transposes.

\subsubsection{{\LayerVariantNameII}}

\begin{figure*}
    \centering
    \includegraphics[width=0.62\textwidth]{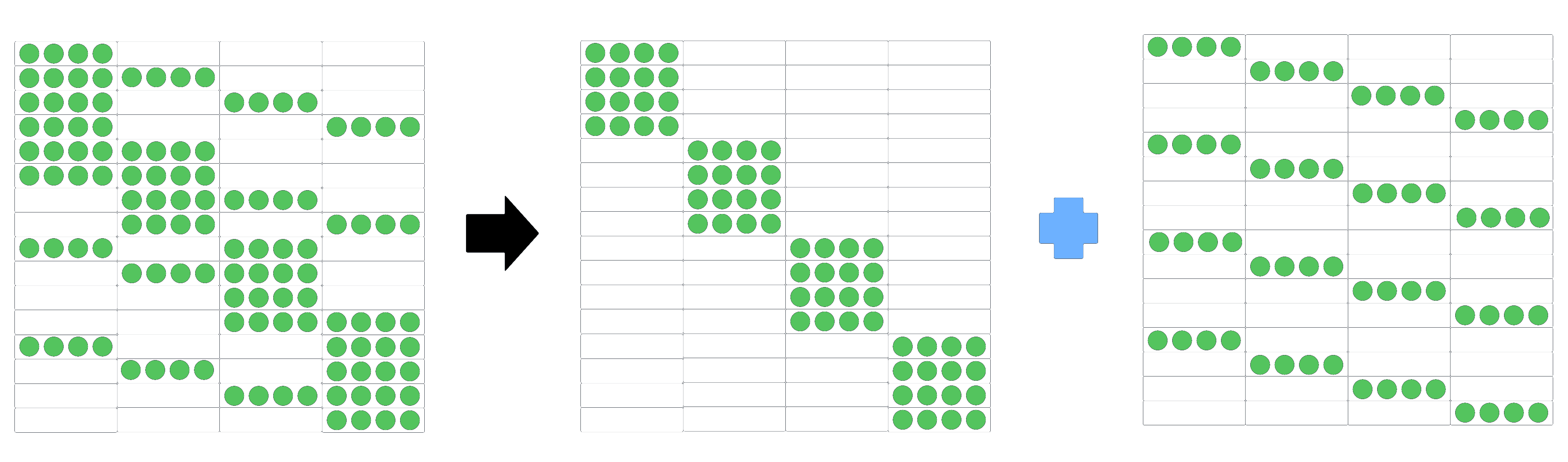}
    \caption{Dyad Output Transposed}
    \label{fig:dyad-ot}
\end{figure*}

The weight matrix and the two split components of {\LayerVariantNameII} is shown in Fig \ref{fig:dyad-ot}. The first component can be calculated exactly the same way as in {\LayerVariantNameI}. The output of the second component can be calculated as $Y_2 = W_2X$. Here, $Y_2$ is the output of the second component, $W_2$ is the weight matrix and $X$ is the activation. Similar to the case of {\LayerVariantNameI}, we can see that if we permute the second component along the rows we can get back a block diagonal matrix. Let the permutation matrix which achieves this be $P$. Since, we are permuting the rows here this permutation matrix needs to be pre multiplied i.e $W_2^{P} = PW_2$ where $W_2^{P}$ is the resultant block diagonal matrix. We can convert $W_2X$ to use this form as shown below.
\begin{align}
    Y_2 &= (P^{T}P)W_2X \\
    Y_2 &= P^{T}(PW_2)X \\
    Y_2 &= P^{T}W_2^{P}X
\end{align}
Here, we can calculate $W_2^{P}X$ similar to the first component and then the permutation by per-multiplying $P^{T}$ can be achieved by transposing the output similar to how it was done for {\LayerVariantNameI}. Thus, similar to {\LayerVariantNameI} we will have a compute complexity of $O({n}_{dyad} \times {n}_{out} \times {n}_{in})$.
\subsubsection{{\LayerVariantNameIII}}
\begin{figure*}
    \centering
    \includegraphics[width=0.62\textwidth]{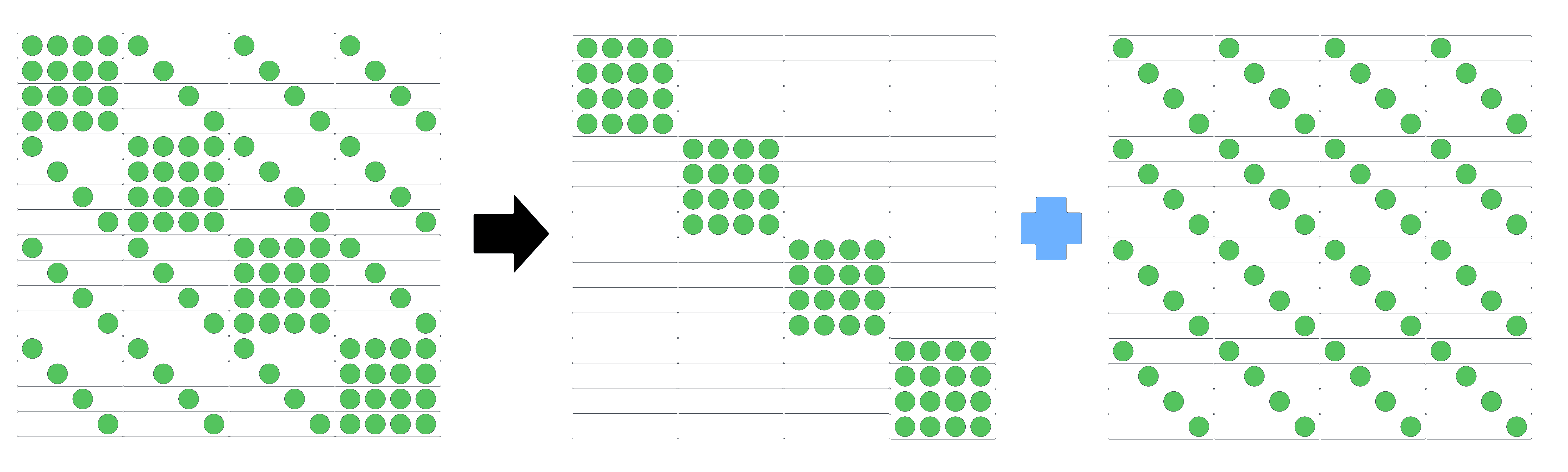}
    \caption{Dyad Double Transposed}
    \label{fig:dyad-dt}
\end{figure*}
Fig \ref{fig:dyad-dt} shows the weight matrix and the components of {\LayerVariantNameIII}. The important thing to note is that the second component can be converted into a block diagonal matrix through a combination of transposing the cloumns as well as transposing the rows. So, in other words it's basically a combination of {\LayerVariantNameI} and {\LayerVariantNameII}. We have to transpose the input before we multiply by the block diagonal weight matrix and then we have to transpose the output to get the final output of the layer.
\begin{align}
    Y_2 &= (P_2^{T}P_2)W_2(P_1P_1^{T})X \\
    Y_2 &= P_2^{T}(P_2W_2P_1)(P_1^{T}X) \\
    Y_2 &= P_2^{T}W_2^{P}X^{'}
\end{align}
The above equations show this. $X^{'} = P_1^{T}X$ is the result of transposing the input while $W_2^{P}$ is the equivalent block diagonal matrix obtained by permuting both the columns and rows ($P_2W_2P_1$). As in the case with the other two variants, this variant also achieves a complexity of $O({n}_{dyad} \times {n}_{out} \times {n}_{in})$.

As further food for thought, we present a sketch discussing some thoughts about the representational power of \textsc{\LayerGroupName} in Appendix \S 5.4.

\section{Experimental Setup: Architectures, Benchmarks and Metrics}
\label{sec:experimental_setup}

\subsection{Choice of Pretraining Corpus}
Since our experiments need multiple pretraining runs to create different pretrained variants of the same architectures, each with the linear layers of the \textit{ff} module replaced by our \textsc{\LayerGroupName} variants, in addition to the baseline \textsc{Dense}, it would be infeasible to pretrain manyfold on full corpora, especially for a new method that can show on-the-fly challenges. Since TinyStories \cite{eldan2023tinystories}, there has been an emerging class of lean pretraining corpora (others being \cite{kaddour2023minipile}, \cite{warstadt2023call}) carefully curated to forsake on superficial aspects of scale (e.g. internet-scale vocab), while being linguistically rich enough. They present a reasonable Goldilocks choice, being small enough to pretrain many runs on, while being large enough to learn emergent LLMesque skills. Hence, we choose \textsc{BabyLM} \cite{warstadt2023call}, which comes in two scales - 100M and 10M tokens respectively. The authors also provide an easy-to-use and ``hackable" setup, with repos that support a) pretraining b) evaluating on \textsc{Blimp}/\textsc{GLUE}. 

\subsection{Models, Architecture, Hyperparameters \& Compute}
We seek a setting which allows direct comparison between \textsc{Dense} vs \textsc{\LayerGroupName}, with preferably simple loss function and minimally randomized training. We avoid encoder-only and encoder-decoder architectures for this reason. To compare with \textsc{BabyLM} baselines, we pick the sole decoder-only architecture they evaluate, i.e. OPT-125m \cite{zhang2022opt}, as the architecture to try our variants with. We lay greater emphasis on exhaustive experiments at 10M data scale, though we also perform a core subset at the 100M scale. To show generalization to higher architecture size, we also repeat some experiments with OPT 350-m. We also present promising results at 10M with Pythia 160-M in Appendix \S5.6.4. We refer to the pretrained \textsc{Dense} checkpoint shared from \textsc{BabyLM} as \textsc{Dense-Ext}, \textsc{Dense} being our replication of it keeping pretraining details same for \textsc{\LayerGroupName}. \textsc{Dyad} variants have $n_{dyad}=4$ unless mentioned ($-8$ i.e. $n_{dyad}=8$). All experiments are on 1 GPU. More compute details are noted in Appendix\S5.5 

\subsection{Benchmarks \& Metrics}
 \label{subsec:benchmarksAndMetrics}
\textbf{Zero-Shot: \textsc{Blimp}} Benchmark of Linguistic Minimal Pairs (\textsc{Blimp}) \cite{warstadt2020blimp} consists of pairs of grammatical-ungrammatical sentences grouped by 12 broad phenomena e.g. anaphora and noun-verb agreement. A good LLM ought to assign higher probability to the grammatical member.

\textbf{Few-Shot: \textsc{OpenLlm}}
The \textsc{OpenLlm} leaderboard \cite{open-llm-leaderboard} has become a prevalent way to benchmark LLMs based on 4 few-shot openbook MCQesque benchmarks. Internally, it uses LMEvalHarness \cite{eval-harness}, which we replicate to compute numbers for our models as well as BabyLm's pretrained checkpoints.


\textbf{Finetuned:\textsc{Glue+}}
General Lang. Understanding Eval (\textsc{Glue}) \cite{wang2018glue}, is a set of 7 NLU tasks, each evaluated post-finetuning. Also, we compute results on \textsc{WSC} and \textsc{BoolQ}. We christen this \textsc{Glue+}.

\textbf{Training Time}
We report both total and \textit{FF}-only (time spent just on \textit{ff} modules) time per minibatch.

\textbf{Memory \& Parameter Footprint}
By storing the dense subset of $W$ as 3D tensor form, \textsc{\LayerGroupName} has lesser space complexity. To gauge real space saved, we measure various notions of memory and parameter size: \\
i) \textbf{Non-Embedding Parameters:} As in Pythia \cite{biderman2023pythia},  we report total Non-Embedding Parameters. \\
ii) \textbf{Model Checkpoint Size:} On-disk size of the model checkpoint. \\
iii) \textbf{In-Training GPU Memory Usage:} During training, models may use memory well beyond parameters, e.g. optimizer state, cached activations etc. In-Training GPU Memory Usage as a metric incorporates this.

\subsection{Results}

\subsubsection{\textsc{\LayerGroupName} vs \textsc{Dense} with 10M tokens}
Through Tables \ref{tab:CrossBenchmarkPerfAllDyadVariants125Cross10Minified} and \ref{tab:CrossBenchmarkPerfAllDyadVariantsPythia160Cross10Minified} (and Appendix Tables 6, 7 \& 8), we see that \textsc{\LayerGroupName} variants are well competitive ($\leq$ 5\%) of the best \textsc{Dense} baseline. 

\begin{figure*}[t]
    \centering
    \includegraphics[width=0.37\textwidth]{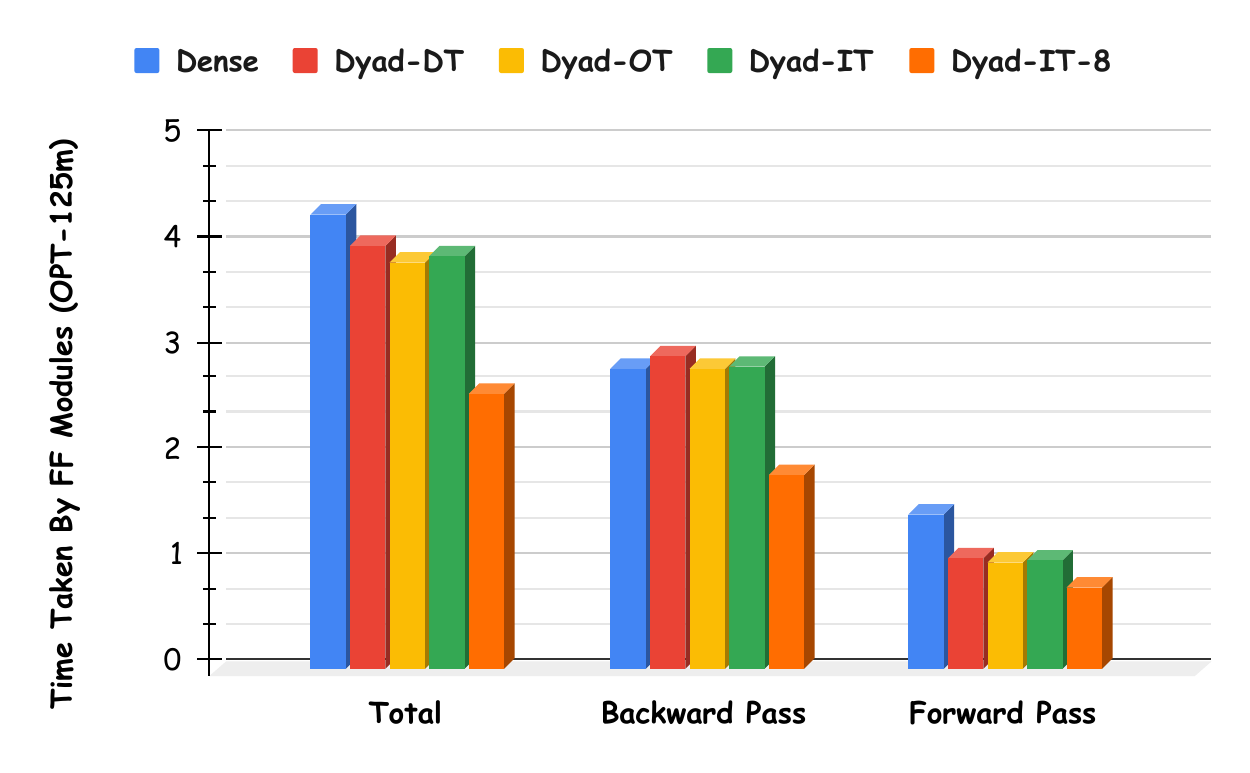}
    \includegraphics[width=0.37\textwidth]{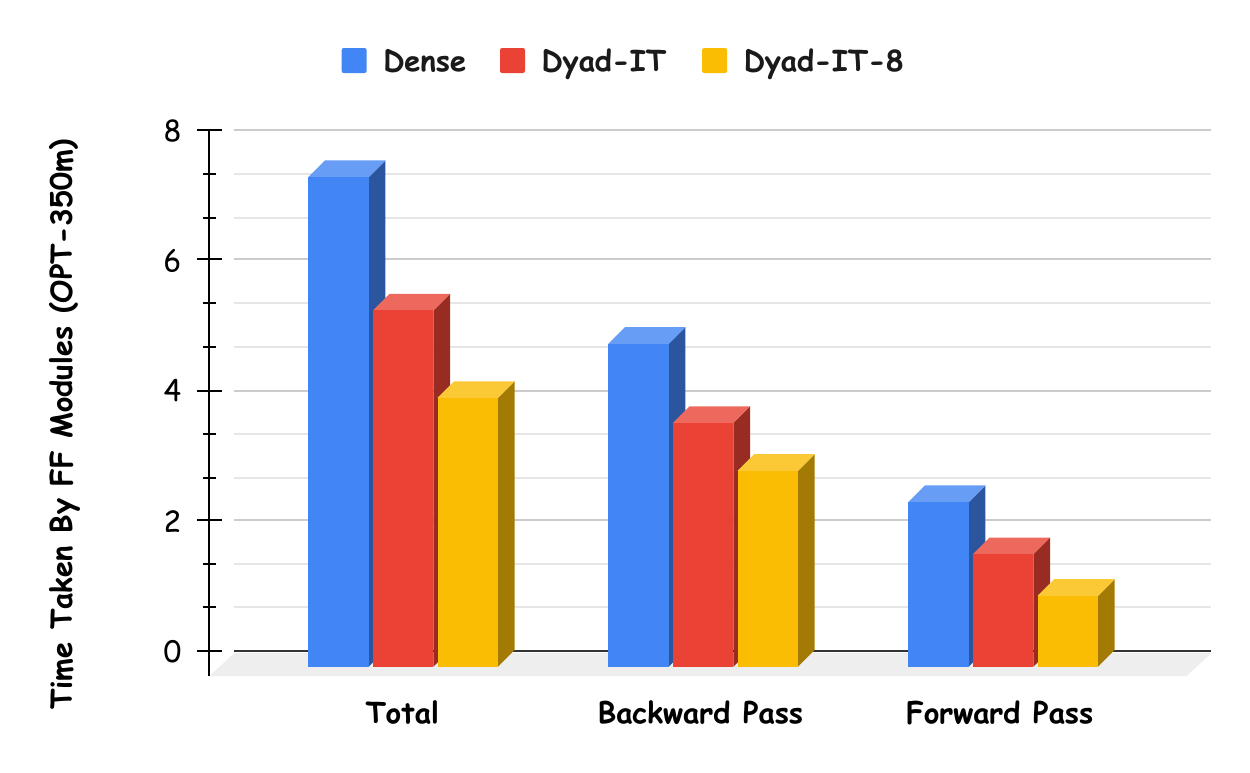}
    \caption{\scriptsize Mean traintime per minibatch by FF modules of OPT-125m/OPT-350m training spent on forward, backward passes and total (Times in ms). \textsc{\LayerGroupName} variants are faster, and $\uparrow$ $n_{dyad}$ (\textsc{\LayerVariantNameI-8}) improves this.} \label{fig:time_both_histograms}
\vspace{-3mm}
\end{figure*}

In addition, through Figures \ref{fig:time_both_histograms},  and Tables \ref{tab:OPT125m_FF_Time}, \ref{tab:Pythia160m_AllModules_Time} and \ref{tab:Pythia160m_FFModules_Time} (as well as Appendix Tables 9 and 10), we see that all \textsc{\LayerGroupName} variants can translate the better complexity to actual speedups. We see that the quantum of these speedups to be much higher for larger architecture sizes i.e. OPT-350m

\subsubsection{Promising Results With Pythia}
\label{subsubsec:promisingResults}
The Pythia suite \cite{biderman2023pythia} of models by EleutherAI, trained based on a permissively licensed collected dataset named The Pile \cite{biderman2022datasheet}.

The results we get by pretraining Pythia on the 10M scale of \textsc{BabyLM} are shared in Table \ref{tab:CrossBenchmarkPerfAllDyadVariantsPythia160Cross10Minified}. We also see that, just as we did for OPT 125-m, the promised time complexity improvements translate into speedups considering both FF-only time (as we can see in Table \ref{tab:Pythia160m_FFModules_Time})  and overall time (as we can see in Table \ref{tab:Pythia160m_AllModules_Time})

\subsubsection{\textsc{*-Cat} experiments}
As can be seen in the \verb|forward()| function of our implementation of \textsc{\LayerVariantNameI} laid out in \S\ref{subsec:PytorchImplementation}, and as we note explicitly therewith (``\textit{We also note that this code \ldots has some overhead \ldots sequential processing of the components}"), having to process the two components underlying our layer, i.e. \textsc{\DyadComponentIShorthand} and \textsc{\DyadComponentIIShorthand} in separate steps does introduce an unnecessary overhead that did not exist for \textsc{Dense}.
To mitigate this, one can conceptualize a slightly faster forward layer that first concatenates the two components to enable parallel processing, before adding them up. We refer to implementations which use this variation by the \textsc{-Cat} suffix, such as in \textsc{\LayerVariantNameI-Cat}. We perform a pretraining run of this variant at \textsc{10M} scale, and find that this is indeed faster as anticipated, while retaining near-identical performance. Specifically, the $ff$-only time per minibatch taken by \textsc{\LayerVariantNameI-Cat} along with OPT-125m is 3.27 ms, rather than 3.90ms taken by the simple, no \textsc{-Cat}, \textsc{\LayerVariantNameI}, which is about 16\% faster. For OPT-350m, the fractional speedup goes up even more, with \textsc{\LayerVariantNameI-Cat} taking 5.46 ms and with \textsc{\LayerVariantNameI} taking 7.92ms, being 45\% faster.

The gains seen by optimizing away even this small overhead point to the promise held by the opportunity to optimize other steps of this layer once matrix multiplication itself has been optimized [through using \textsc{\LayerGroupName} style layers].

\subsubsection{Profiling Experiments At Wider Architectural Scales} \label{subsubsec:ProfilingWideArch}
Since \textsc{Dyad} is primarily applied herein to \textit{ff} module, assessing its benefits at higher relative width would give us important additional insight on its salience and generalizability in terms of benefit.

To do this, we take the OPT-1.3B model's architecture but cap its depth down to 6 layers so that the model continue to fit within our computational constraints at levels of width all the way upto 4096.

\subsubsection{Testing Waters with Vision Applicability - MNIST Experiment(s)} \label{subsubsec:MNISTExperiments}
Since the bulk of our experiments as well as intuitions and writing is in a large language model/NLP context, a natural question that may perplex a reader is if using a \textsc{\LayerGroupName} style linear layer rather than a \textsc{Dense} one holds promise in other modalities e.g. computer vision. To make a basic probe in this direction, we do experiments with the simple but foundational MNIST digit classification task \cite{lecun1998mnist}, replacing linear layers with both their plain \textsc{Dense} and our \textsc{\LayerVariantNameI}, again with $n_{dyad}=4$. Furthermore, we also test the waters in terms of trying our approaches with diverse accelerators by performing these experiments directly on a Macbook CPU without using a GPU or MPS etc. 

We find the properties of reasonable performance preservation and speedups in both ff and overall time carry over to this situation too. Specifically, we find \textsc{\LayerVariantNameI} achieves 98.51\% test accuracy vs the 98.43\% achieved by \textsc{Dense}, while taking 3.76 seconds of $ff$-only time per minibatch compared to 4.85 seconds by \textsc{Dense}.

\begin{figure*}[!h]
    \centering
    \includegraphics[width=0.83\textwidth]{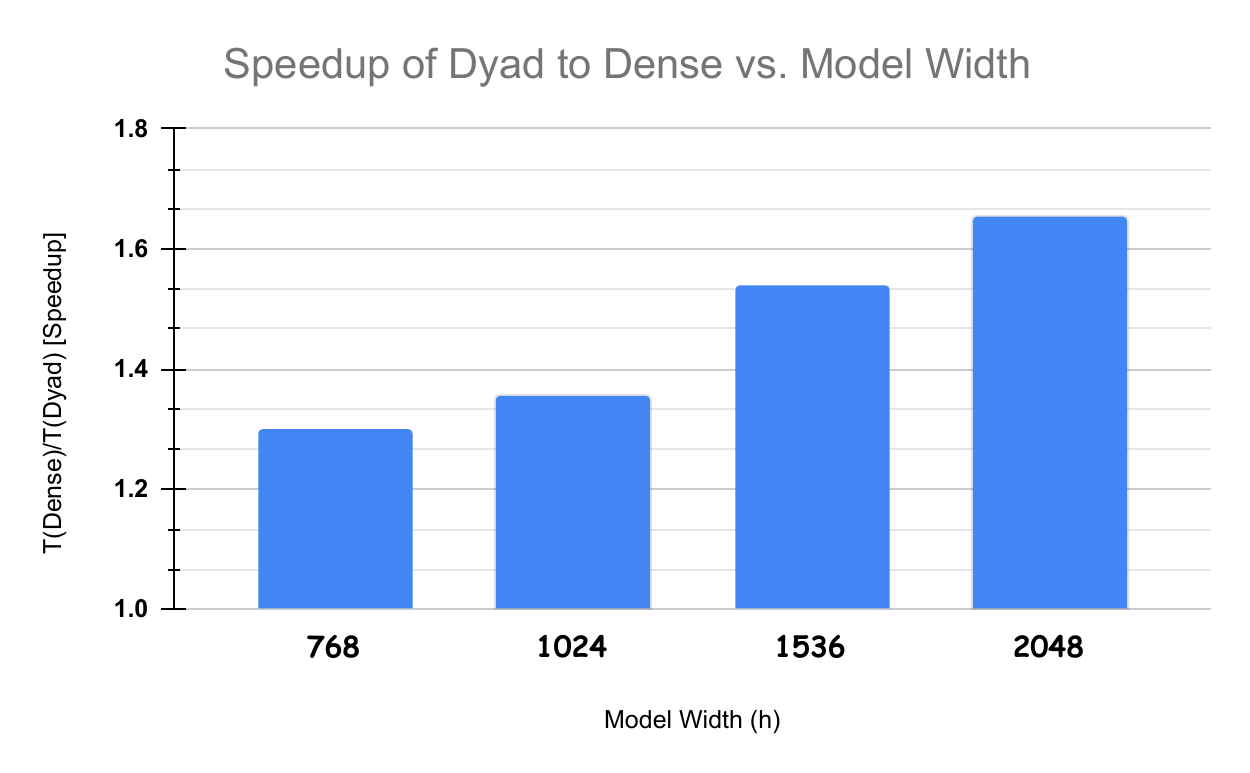}
    \vspace{-3mm}
    \caption{\small \textsc{\LayerGroupName} vs \textsc{Dense} Speedup At Different Model Widths of 6-Layer Capped OPT-like architecture.} \label{fig:DyadVsDenseSpeedup}
\vspace{-3mm}
\end{figure*}

\begin{table}[!ht]
    \centering
    \begin{tabular}{|l|l|l|l|l|}
    \hline
        Model & Forward Pass & Backward Pass & Total & Total speedup ratio \\ \hline
        \textsc{Dense} & 1.458818136 & 2.843522568 & 4.302340703 & 1 \\ \hline
        \textsc{\LayerVariantNameI} & 1.037282137 & 2.864683089 & 3.901965226 & 1.102608674 \\ \hline
        \textsc{\LayerVariantNameII} & 1.005873492 & 2.833987413 & 3.839860905 & 1.12044181 \\ \hline
        \textsc{\LayerVariantNameIII} & 1.048527787 & 2.955974824 & 4.004502611 & 1.074375802 \\ \hline
        \textsc{\LayerVariantNameI}-8 & 0.7726907735 & 1.836098994 & 2.608789767 & 1.649171105 \\ \hline
    \end{tabular}
    \smallskip
    \caption{\footnotesize Mean time taken per minibatch by the ff transformer modules of OPT-125m training on account of forward, backward passes and in total. All times are in milliseconds. Speedup ratio is computed w.r.t. \textsc{Dense}}
    \label{tab:OPT125m_FF_Time}
\end{table}

\begin{figure*}[t]
    \centering
    \includegraphics[width=0.37\textwidth]{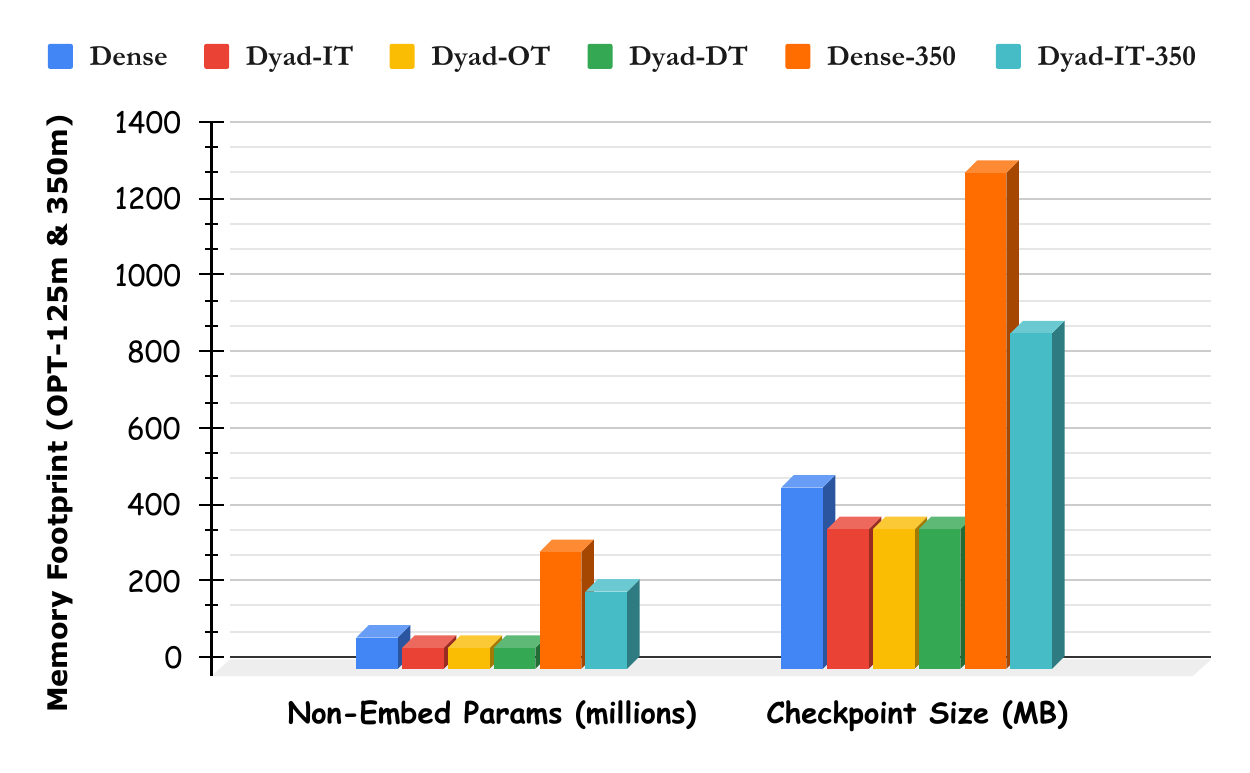}
    \includegraphics[width=0.37\textwidth]{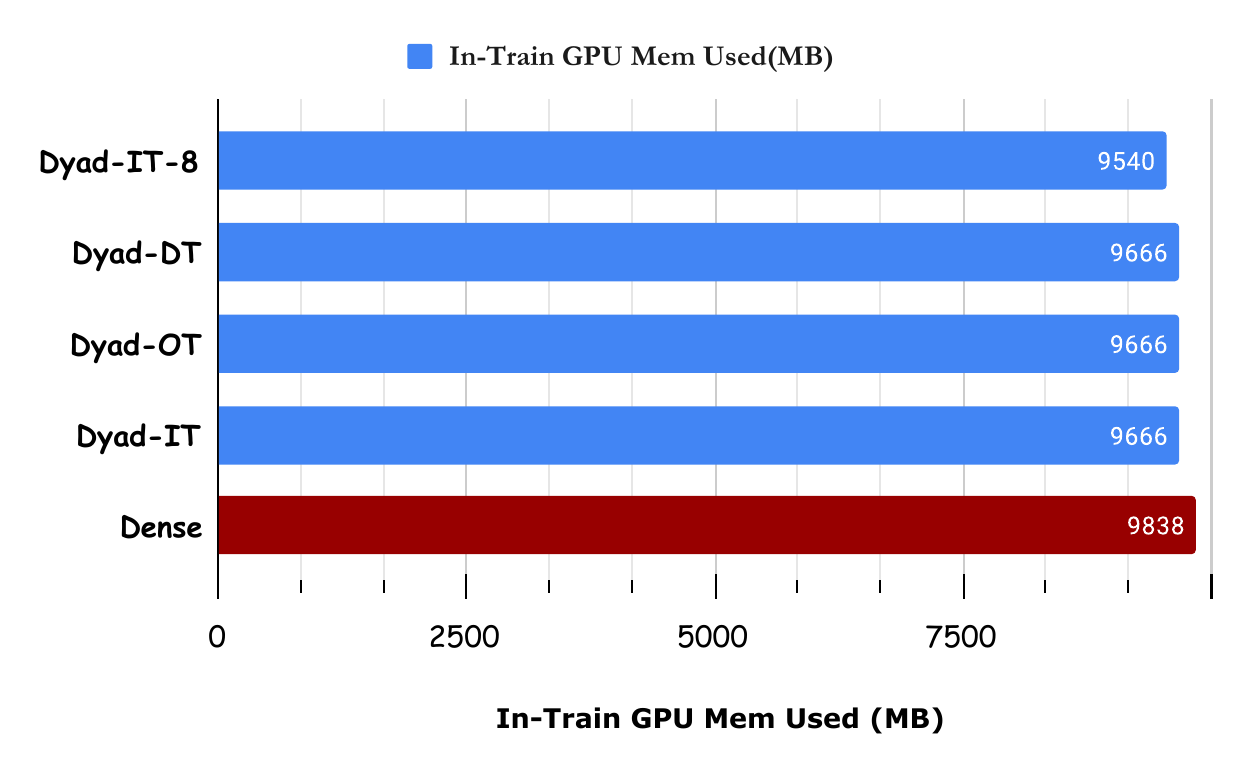}
    \caption{\scriptsize Memory and parameter footprint of OPT-125m/OPT-350m training as per various static estimates on the left and dynamic GPU mem usage on the right.} \label{fig:space_both_histograms}
\vspace{-3mm}
\end{figure*}

\nprounddigits{2} 
\begin{table*}[!t]
        \centering \tiny
        \begin{tabular}{| c | c | c | c | c | c | c | c |} \hline
   Benchmark & Task & \textsc{Dense} & \textsc{Dense-Ext} & \textsc{\LayerVariantNameI} & \textsc{\LayerVariantNameII} & \textsc{\LayerVariantNameIII} & \textsc{\LayerVariantNameI}-8\\ \hline \hline
 \multirow{3}{*}{\textsc{GLUE+} Mean}   & GLUE+	& 68.82	& 63.38	& \underline{67.33}	& \underline{68.46}	& \underline{68.59}	& \underline{67.70} \\ 
& GLUE+-QA	& 66.37	& 63.67 & \underline{66.27}	& \underline{66.27} & \underline{63.69}	& \underline{64.02} \\ 
& GLUE+-NLI	& 68.27 & 59.78 & \underline{65.64} & \underline{68.27} & \underline{\textbf{68.67}} & \underline{67.65}       
        \\ \hline \hline
         \multirow{1}{*}{\textsc{Blimp} Mean}   &  \textsc{Blimp} & 59.16 & 60.31	& \underline{\textbf{60.47}} & \underline{\textbf{62.55}} & \underline{\textbf{60.86}} & 58.88
        \\ \hline \hline
  \multirow{1}{*}{\textsc{OpenLlm} Means}   &    \textsc{OpenLlm} & 30.27	& 30.39	& \underline{\textbf{30.61}}	& \underline{\textbf{30.74}} &	\underline{\textbf{30.58}} & \underline{\textbf{30.65}}
        \\ \hline
        \end{tabular}
        \vspace{-0.6\abovedisplayskip}
        \caption{\footnotesize Performance on \textsc{GLUE+} (finetuning), \textsc{Blimp} (0-shot), \textsc{OpenLlm} (few-shot) benchmarks for \textsc{Dense} baselines vs 3 \textsc{\LayerGroupName} variants with $n_{dyad}=4$ and a sparser version of the 1st (\textsc{\LayerVariantNameI}-8). Numbers which exceed \textsc{Dense}/\textsc{Dense-Ext} are bolded/underlined respectively. All \textsc{\LayerGroupName} variants are $\geq 0.95\times{}max(\textsc{Dense},\textsc{Dense-Ext})$. We present aggregates for brevity and defer individual values to Appendix Table 2}
        \label{tab:CrossBenchmarkPerfAllDyadVariants125Cross10Minified}
\vspace{-3ex}
\end{table*}
\nprounddigits{0}

\begin{table*}[!ht]
        \centering \tiny
        \begin{tabular}{| c |c | c | c|} \hline
        Benchmark & Task & \textsc{Dense}  & \textsc{\LayerVariantNameI}\\ \hline \hline
 \multirow{3}{*}{Means}       & \textsc{GLUE+}	     & 73.86818182 & \textbf{73.71942857} \\ 
      & \textsc{GLUE+-QA}  & 69.875 & \textbf{72.7185} \\ 
      & \textsc{GLUE+-NLI}  & 73.188 & \textbf{77.2675} \\ 
        \hline \hline
 \multirow{1}{*}{Means}   &  \textsc{Blimp}	& 58.87623529 & \textbf{59.26882353} \\
       \hline \hline
   \multirow{1}{*}{Means}   &    \textsc{OpenLlm} & 29.9997 & \textbf{30.05875}
        \\ \hline
        \end{tabular}
        \caption{\small Benchmark numbers for Pythia-160m pretrained at the \textsc{10M} scale comparing \textsc{Dense} with \textsc{\LayerVariantNameI}. Instances where \textsc{\LayerVariantNameI} exceeds \textsc{Dense} are marked in bold, while instances where \textsc{\LayerVariantNameI} falls below 0.95* \textsc{Dense} are marked in {\color{red} Red}. \textsc{\LayerVariantNameI} falls below the 0.95\% mark w.r.t. \textsc{Dense} on only 3 zero-shot and 2 \textbf{GLUE+} tasks, falling above the mark  on all \textsc{GLUE+} aggregate tasks and \textsc{OpenLlm}. We present aggregates for brevity and defer individual values to Appendix Table 7}
        \label{tab:CrossBenchmarkPerfAllDyadVariantsPythia160Cross10Minified}
\end{table*}

\begin{table}[!ht]
    \centering
    \begin{tabular}{|c|l|l|l|l|}
    \hline
        Model & Forward Pass & Backward Pass & Total & Total speedup ratio \\ \hline
        Dense & 101.89 & 220.16 & 332.64 & 1 \\ \hline
        \textsc{\LayerVariantNameI} & 89.40 & 229.86 & 310.62 & 1.071 \\ \hline
    \end{tabular}{}
    \smallskip
    \caption{\small Mean time taken per minibatch by all modules of Pythia-160m training on account of forward, backward passes and in total. Times are in milliseconds. Speedup ratio is computed w.r.t. \textsc{Dense}}
    \label{tab:Pythia160m_AllModules_Time}
\end{table}

\begin{table}[!ht]
    \centering
    \begin{tabular}{|c|l|l|l|l|}
    \hline
        Model & Forward Pass & Backward Pass & Total & Total speedup ratio \\ \hline
        Dense & 1.414 & 2.826 & 4.240 & 1 \\ \hline
        \textsc{\LayerVariantNameI} & 1.070 & 2.879 & 3.949 & 1.074 \\ \hline
        \textsc{\LayerVariantNameI-8} & 0.795 & 1.843 & 2.637 & 1.607 \\ \hline
    \end{tabular}{}
    \smallskip
    \caption{\small Mean time taken per minibatch by the $ff$ (feedforward) modules of Pythia-160m training on account of forward, backward passes and in total. All times are in milliseconds. Speedup ratio is computed w.r.t. \textsc{Dense}}
    \label{tab:Pythia160m_FFModules_Time}
\end{table}


\section{Future Work}
\label{sec:conclusion_future_work}

In the future, we aim to explore i) using a heterogeneous mix of \textsc{\LayerGroupName} variants to approximate different ff layers ii) Replicating our experiments other minified corpora such as Minipile \cite{kaddour2023minipile}.

\textbf{Acknowledgements:}
We thank Ryan McDonald and Nirmal Mukhi (ASAPP Inc.), the workshop organizers of the WANT and ESNLP workshops as well as anonymous reviewers for their helpful feedback.

\clearpage

\bibliographystyle{neurips}
\bibliography{dyadPaperCitations}

\begin{thebibliography}{10}

\bibitem{krizhevsky2012imagenet}
Krizhevsky, A., I.~Sutskever, G.~E. Hinton.
\newblock Imagenet classification with deep convolutional neural networks.
\newblock \emph{Advances in neural information processing systems}, 25, 2012.

\bibitem{vaswani2017attention}
Vaswani, A., N.~Shazeer, N.~Parmar, et~al.
\newblock Attention is all you need.
\newblock \emph{Advances in neural information processing systems}, 30, 2017.

\bibitem{sutskever2014sequence}
Sutskever, I., O.~Vinyals, Q.~V. Le.
\newblock Sequence to sequence learning with neural networks.
\newblock \emph{Advances in neural information processing systems}, 27, 2014.

\bibitem{srinivas2021bottleneck}
Srinivas, A., T.-Y. Lin, N.~Parmar, et~al.
\newblock Bottleneck transformers for visual recognition.
\newblock In \emph{Proceedings of the IEEE/CVF conference on computer vision and pattern recognition}, pages 16519--16529. 2021.

\bibitem{devlin2018bert}
Devlin, J., M.-W. Chang, K.~Lee, et~al.
\newblock Bert: Pre-training of deep bidirectional transformers for language understanding.
\newblock \emph{arXiv preprint arXiv:1810.04805}, 2018.

\bibitem{raffel2020exploring}
Raffel, C., N.~Shazeer, A.~Roberts, et~al.
\newblock Exploring the limits of transfer learning with a unified text-to-text transformer.
\newblock \emph{The Journal of Machine Learning Research}, 21(1):5485--5551, 2020.

\bibitem{dettmers2023qlora}
Dettmers, T., A.~Pagnoni, A.~Holtzman, et~al.
\newblock Qlora: Efficient finetuning of quantized llms.
\newblock \emph{arXiv preprint arXiv:2305.14314}, 2023.

\bibitem{kitaev2020reformer}
Kitaev, N., {\L}.~Kaiser, A.~Levskaya.
\newblock Reformer: The efficient transformer.
\newblock \emph{arXiv preprint arXiv:2001.04451}, 2020.

\bibitem{qin2022cosformer}
Qin, Z., W.~Sun, H.~Deng, et~al.
\newblock cosformer: Rethinking softmax in attention.
\newblock \emph{arXiv preprint arXiv:2202.08791}, 2022.

\bibitem{dao2022flashattention}
Dao, T., D.~Fu, S.~Ermon, et~al.
\newblock Flashattention: Fast and memory-efficient exact attention with io-awareness.
\newblock \emph{Advances in Neural Information Processing Systems}, 35:16344--16359, 2022.

\bibitem{wiki-matrix-perm-properties}
Wikipedia, W.~contributing authors.
\newblock Permutation matrix: Properties.
\newblock \url{https://en.wikipedia.org/wiki/Permutation_matrix#Properties}, 2023.

\bibitem{eldan2023tinystories}
Eldan, R., Y.~Li.
\newblock Tinystories: How small can language models be and still speak coherent english?
\newblock \emph{arXiv preprint arXiv:2305.07759}, 2023.

\bibitem{kaddour2023minipile}
Kaddour, J.
\newblock The minipile challenge for data-efficient language models.
\newblock \emph{arXiv preprint arXiv:2304.08442}, 2023.

\bibitem{warstadt2023call}
Warstadt, A., L.~Choshen, A.~Mueller, et~al.
\newblock Call for papers--the babylm challenge: Sample-efficient pretraining on a developmentally plausible corpus.
\newblock \emph{arXiv preprint arXiv:2301.11796}, 2023.

\bibitem{zhang2022opt}
Zhang, S., S.~Roller, N.~Goyal, et~al.
\newblock Opt: Open pre-trained transformer language models.
\newblock \emph{arXiv preprint arXiv:2205.01068}, 2022.

\bibitem{warstadt2020blimp}
Warstadt, A., A.~Parrish, H.~Liu, et~al.
\newblock Blimp: The benchmark of linguistic minimal pairs for english.
\newblock \emph{Transactions of the Association for Computational Linguistics}, 8:377--392, 2020.

\bibitem{open-llm-leaderboard}
Beeching, E., C.~Fourrier, N.~Habib, et~al.
\newblock Open llm leaderboard.
\newblock \url{https://huggingface.co/spaces/HuggingFaceH4/open_llm_leaderboard}, 2023.

\bibitem{eval-harness}
Gao, L., J.~Tow, S.~Biderman, et~al.
\newblock A framework for few-shot language model evaluation, 2021.

\bibitem{wang2018glue}
Wang, A., A.~Singh, J.~Michael, et~al.
\newblock Glue: A multi-task benchmark and analysis platform for natural language understanding.
\newblock In \emph{Proceedings of the 2018 EMNLP Workshop BlackboxNLP: Analyzing and Interpreting Neural Networks for NLP}, pages 353--355. 2018.

\bibitem{biderman2023pythia}
Biderman, S., H.~Schoelkopf, Q.~G. Anthony, et~al.
\newblock Pythia: A suite for analyzing large language models across training and scaling.
\newblock In \emph{International Conference on Machine Learning}, pages 2397--2430. PMLR, 2023.

\bibitem{biderman2022datasheet}
Biderman, S., K.~Bicheno, L.~Gao.
\newblock Datasheet for the pile.
\newblock \emph{arXiv preprint arXiv:2201.07311}, 2022.

\bibitem{lecun1998mnist}
LeCun, Y.
\newblock The mnist database of handwritten digits.
\newblock \emph{http://yann. lecun. com/exdb/mnist/}, 1998.

\end{thebibliography}
\clearpage

\section{Appendix}
\label{sec:appendix}

\subsection{Important Additional Caveats About Formulation \& Implementation}

\begin{enumerate}
    \item \textbf{Constraints on Rectangular $W$ dimensions:} Since $n_{dyad}$ denotes the number of equi-sized blocks (with W's dimensions being factorable out at $n_{in} \times n_{dyad}$ and $n_{out}\times n_{dyad}$, for a non-trivial sparse reduction, the dimensions of W woud need to be both divisible by some $n_{dyad}>1$ --- one cannot divide a $7 \times 6$ matrix into $4 \times 4$ blocks. However, we can see that for practical usage this aspect is somewhat pedantic , one can always pad up the dimensions with zeroes to different extents such that $n_{dyad}$, i.e., the desired level of sparsity is attained - e.g, in the $7 \times 6$ case, zero-padding up the number of rows by 1, our dimensions will now have a common factor $2$,
    \item \textbf{Additional Kernel Launches in Implementation:} The code for \textsc{\LayerVariantNameI} described in the Formulation section does have some overhead in terms of additional kernel launches but for larger sized models this overhead will amortize away.
\end{enumerate}

\subsection{Hyperparameter Choices \& Compute Details}
For simplicity, we avoid mixed precision training (use fp32 throughout), gradient checkpointing or quantization. Since \textsc{BabyLM}'s training setup required using earlier versions of Pytorch than would be compatible don't use FlashAttention. These techniques are in either case not intertwined directly to our method. All our OPT-125m experiments for both the \textsc{Strict} and \textsc{StrSma} scales were done on a NVIDIA V100. For OPT-350m experiments, we use a A10G.

We use the optimizers for dense layers from the typical \textsc{Dense} setting as-is, sans any changes particular to our method. Initialization for the various \textsc{Sparse} approaches too, was done in the same fashion as for \textsc{Dense}. Optimizers same

\subsection{Additional Results}

\subsubsection{Complete Benchmark Result Tables}

\nprounddigits{2} 
\begin{table*}[!t]
        \centering \tiny
        \begin{tabular}{| c |c | n{2}{2} | n{2}{2} | n{2}{2} | n{2}{2} | n{2}{2} | n{2}{2} |} \hline
   Benchmark & Task & \textsc{Dense} & \textsc{Dense-Ext} & \textsc{\LayerVariantNameI} & \textsc{\LayerVariantNameII} & \textsc{\LayerVariantNameIII} & \textsc{\LayerVariantNameI}-8\\ \hline \hline
  \multirow{11}{*}{\textsc{GLUE+}}  &    CoLA & 68.499 & 64.6 & 68.204 & 68.106 & 67.321 & 67.419 \\ 
        & SST-2 & 86.417 & 81.9 & 86.614 & 85.827 & 85.039  & 85.236\\  
        & MPRC (F1) & 76.562 & 72.5 & 77.444 & 76.981 & 78.491 & 73.563\\ 
        & QQP (F1) & 80.496 & 60.4 & 79.79 & 80.26 & 80.909 & 80.854\\ 
        & MNLI & 70.771 & 57.6 & 71.122 & 71.32 & 70.893 & 70.817\\  
        & MNLI-mm & 71.801 & 60.0 & 72.056 & 72.518 & 72.565 & 70.99\\  
        & QNLI & 69.904 & 61.5 & 70.91 & 76.728 & 74.672 & 70.21\\  
        & RTE & 60.606 & 60.0 & 48.485 & 52.525 & 56.566 & 58.586\\  
        & BoolQ & 66.252 & 63.3 & 64.315 & 63.9 & 63.624 & 64.177\\  
        & MultiRC & 56.298 & 55.2 & 57.065 & 57.941 & 48.959 & 54.326\\  
        & WSC & 49.398 & 60.2 & 44.578 & 46.988 & 55.422 & 48.193\\ \hline
 \multirow{3}{*}{\textsc{GLUE+} Means}   & GLUE+	& 68.81854545	& 63.38181818	& 67.32572727	& 68.46309091	& 68.58736364	& 67.6952 \\ 
& GLUE+-QA	& 66.37066667	& 63.66666667 & 66.27466667	& 66.274 & 63.69133333	& 64.022 \\ 
& GLUE+-NLI	& 68.2705 & 59.775 & 65.64325 & 68.27275 & 68.674 &67.65075       
        \\ \hline \hline
    \multirow{17}{*}{\textsc{Blimp}}    & Anaphor Agr. & 49.489 & 63.8 & 67.331 & 64.877 & 73.926 & 59.254\\  
        & Agr. Structure & 68.101 & 70.6 & 71.339 & 68.465 & 68.647 & 67.823\\  
        & Binding & 68.67 & 67.1 & 65.954 & 65.183 & 63.595 & 68.937\\ 
        & Control/Raising & 66.637 & 66.5 & 63.522 & 64.273 & 63.831 & 62.726\\  
        & D-N Agr. & 74.198 & 78.5 & 81.053 & 81.252 & 80.151 & 74.251\\  
        & Ellipsis & 57.333 & 62 & 61.085 & 57.506 & 54.215 & 55.081\\  
        & Filler-Gap & 65.359 & 63.8 & 64.581 & 65.64 & 66.667 & 65.951\\  
        & Irregular Forms & 77.659 & 67.5 & 82.748 & 75.878 & 81.781 & 66.616\\  
        & Island Effects & 44.021 & 48.6 & 54.746 & 49.888 & 47.347 & 48.281\\  
        & NPI Licensing & 41.193 & 46.7 & 47.464 & 42.955 & 49.59 & 39.311\\  
        & Quantifiers & 61.566 & 59.6 & 53.658 & 71.458 & 44.874 & 67.13\\  
        & Subject-Verb Agreement & 54.616 & 56.9 & 55.772 & 61.12 & 56.947 & 63.884\\  
        & Hypernym & 49.186 & 50.0 & 48.721 & 46.744 & 49.302 & 50.698\\ 
        & QA Congruence (Easy) & 57.812 & 54.7 & 59.375 & 60.938 & 54.688 & 57.812\\  
        & QA Congruence (Tricky) & 32.727 & 31.5 & \boldmath 35.758 & 47.879 & 39.394 & 39.394\\  
        & Subject Auxiliary Inversion & 73.92 & 80.3 & 56.014 & 70.773 & 72.92 & {73.506}\\  
        & Turn Taking & 63.214 & 57.1 & 58.929 & 68.571 & 66.786 & 55.0\\ \hline
         \multirow{1}{*}{Means}   &  \textsc{Blimp} & 59.15888235 & 60.30588235	& 60.47352941 & 62.55294118 & 60.86241176 & 58.8843125
        \\ \hline \hline
    \multirow{4}{*}{\textsc{OpenLlM}}     & ArcChallenge-25 & 22.781 & 23.720 & 22.867 & 25.256 & 23.293 & {23.293} \\  
        & Hellaswag-10 & 25.811 & 25.114 & 25.164 & 24.766 & 24.796 & 25.433 \\  
        & TruthfulQA-MC-0 & 49.393 & 49.720 & 51.122  & 48.828 & 49.843 & 49.681\\  
        & MMLU-5 & 23.11 & 23.007 & 23.304  & 24.097  & 24.397 & 24.199 \\ \hline
  \multirow{1}{*}{Means}   &    \textsc{OpenLlm} & 30.27375	& 30.39025	& 30.61425	& 30.73675 &	30.58225 &	30.6515 
        \\ \hline
        \end{tabular}
        \caption{Performance on \textsc{GLUE+} (post-finetuning), \textsc{Blimp} (zero-shot), \textsc{OpenLlm} (few-shot) benchmarks for the \textsc{Dense} and \textsc{Dense-Ext} baselines and all 3 {\LayerGroupName} variants as well as a doubly sparser version of the 1st variant. These results are with OPT-125m when pretrained at the 10M scale, a summary of which is presented in the results - the rows corresponding to Benchmark aggregate means from this table were presented in Table 1 of the main paper.}
        \label{tab:CrossBenchmarkPerfAllDyadVariants125Cross10Magisterial}
\end{table*}
\nprounddigits{0}

\begin{table*}[!t]
        \centering \tiny
        \begin{tabular}{| c |c | c | c|} \hline
        Benchmark & Task & \textsc{Dense}  & \textsc{\LayerVariantNameI}\\ \hline \hline
  \multirow{11}{*}{\textsc{GLUE+}}  &   CoLA & 70.069  & 69.48\\  
        & SST-2 & 85.039  & 85.039\\  
        & MPRC (F1) & 80.435  & 79.715\\  
        & QQP (F1) & 81.125  & 81.356\\  
        & MNLI & 71.853  & 70.908\\ 
        & MNLI-mm & 73.297  & 71.929\\  
        & QNLI & 80.315  & 76.859\\  
        & RTE & 53.535  & {\color {red} 43.434}\\  
        & BoolQ & 64.73  & 64.315\\ 
        & MultiRC & 49.726  & 50.383\\  
        & WSC & 53.012  & \textbf{59.036}\\ \hline
 \multirow{3}{*}{Means}       & \textsc{GLUE+}	& 69.376	& \textbf{72.220} \\
        & \textsc{GLUE+-QA}	& 64.964	& 64.804 \\
        & \textsc{GLUE+-NLI}	& 69.750	& \textbf{73.232} \\
        \hline \hline
\multirow{17}{*}{\textsc{Blimp}}    & Anaphor Agr. & 62.168  & 60.685\\ 
        & Agr. Structure & 69.241  & 67.083\\ 
        & Binding & 72.069 & {\color {red} 66.014}\\  
        & Control/Raising & 67.852  & {\color {red} 61.6}\\  
        & D-N Agr. & 87.019  & 84.633\\  
        & Ellipsis & 62.875  & 63.279\\  
        & Filler-Gap & 68.830  & 68.659\\ 
        & Irregular Forms & 84.173  & {\color {red} 73.232}\\  
        & Island Effects & 46.375  & \textbf{52.242}\\  
        & NPI Licensing & 57.060 & {\color{red} 46.083}\\  
        & Quantifiers & 68.959  & 66.718\\  
        & Subject Verb Agreement & 67.66  & {\color{red} 59.422}\\ 
        & Hypernym & 44.651  & \textbf{50}\\ 
        & QA Congruence (Easy) & 54.688  & {\color{red} 50}\\  
        & QA Congruence (Tricky) & 47.879  & \textbf{50.303}\\  
        & Subject Auxiliary Inversion & 78.970  & {\color{red} 64.089}\\  
       &  Turn Taking & 64.286  & 61.071\\ \hline
 \multirow{1}{*}{Means}   &  \textsc{Blimp}	& 64.98 & {\color{red} 61.47} \\
       \hline \hline
    \multirow{4}{*}{\textsc{OpenLlm}}     & ArcChallenge-25 & 23.379  & 24.500   \\  
        & Hellaswag-10 & 25.085  & 25.035   \\  
        & TruthfulQA-MC-0 & 48.661   & 50.291   \\  
        & MMLU-5 & 23.190   &  22.910  \\ \hline
   \multirow{1}{*}{Means}   &    \textsc{OpenLlm} & 30.078 & 30.680
        \\ \hline
        \end{tabular}
        \caption{Benchmark numbers for OPT-350m pretrained at the \textsc{10M} scale comparing \textsc{Dense} with \textsc{\LayerVariantNameI}. Instances where \textsc{\LayerVariantNameI} exceeds \textsc{Dense} are marked in bold, while instances where \textsc{\LayerVariantNameI} falls below 0.95* \textsc{Dense} are marked in {\color{red} Red}. We can see this happens only for four zero-shot tasks and none of the few-shot tasks.}
        \label{tab:CrossBenchmarkPerfAllDyadVariants350Cross10Magisterial}
\end{table*} 

\begin{table*}[!t]
        \centering \tiny
        \begin{tabular}{| c |c | c | c|} \hline
        Benchmark & Task & \textsc{Dense}  & \textsc{\LayerVariantNameI}\\ \hline \hline
  \multirow{11}{*}{\textsc{GLUE+}}  &   CoLA & 68.4 & \textbf{68.597} \\ \hline 
        & SST-2 & 85.236 & 84.843\\  
        & MPRC (F1) & 78.873 & 78.261\\  
        & QQP (F1) & 80.336 & 80.54\\  
        & MNLI & 70.451 & 69.918\\  
        & MNLI-mm & 70.321 & \textbf{70.974}\\  
        & QNLI & 55.118 & \textbf{73.447} \\  
        & RTE & 48.485 & {\color{red} 44.444}\\ 
        & BoolQ & 66.113 & 65.422\\  
        & MultiRC & 55.75 & {\color{red} 51.698}\\  
        & WSC & 42.169 & \textbf{53.012}\\  \hline
 \multirow{3}{*}{Means}       & \textsc{GLUE+}	     & 73.86818182 & \textbf{73.71942857} \\ 
      & \textsc{GLUE+-QA}  & 69.875 & \textbf{72.7185} \\ 
      & \textsc{GLUE+-NLI}  & 73.188 & \textbf{77.2675} \\ 
        \hline \hline
\multirow{17}{*}{\textsc{Blimp}}          &  Anaphor Agr. & 56.851 & 55.419\\ 
       & Agr. Structure & 68.671 & \textbf{68.708} \\  
       & Binding & 67.854 & 64.619\\  
       & Control/Raising & 58.948 & \textbf{59.677}\\  
       & D-N Agr. & 75.55 & \textbf{75.961}\\  
       &  Ellipsis & 62.356 & 60.624\\  
       & Filler-Gap & 59.835 & \textbf{61.656}\\  
       & Irregular Forms & 56.132 & \textbf{57.30}3\\  
       & Island Effects & 51.345 & \textbf{53.812} \\  
       & NPI Licensing & 48.558 & \textbf{55.421} \\  
       & Quantifiers & 60.768 & 58.733\\  
       & Subject Verb Agreement & 58.229 & {\color{red} 53.64}\\  
       & Hypernym & 49.07 & \textbf{52.791}\\  
       & QA Congruence (Easy) & 53.125 & \textbf{57.812}\\  
       & QA Congruence (Tricky) & 39.394 & \textbf{51.515}\\  
       & Subject Auxiliary Inversion & 68.139 & {\color{red} 61.308}\\  
       & Turn Taking & 66.071 & {\color{red} 58.571}\\ \hline 
 \multirow{1}{*}{Means}   &  \textsc{Blimp}	& 58.87623529 & \textbf{59.26882353} \\
       \hline \hline
    \multirow{4}{*}{\textsc{OpenLlm}}     & ArcChallenge-25 & 23.549  & 22.696   \\  
        & Hellaswag-10 & 26.260  &  25.423 \\  
        & TruthfulQA-MC-0 & 46.972   & \textbf{48.618}   \\  
        & MMLU-5 & 23.218   &  \textbf{23.498}  \\ \hline
   \multirow{1}{*}{Means}   &    \textsc{OpenLlm} & 29.9997 & \textbf{30.05875}
        \\ \hline
        \end{tabular}
        \caption{Benchmark numbers for Pythia-160m pretrained at the \textsc{10M} scale comparing \textsc{Dense} with \textsc{\LayerVariantNameI}. Instances where \textsc{\LayerVariantNameI} exceeds \textsc{Dense} are marked in bold, while instances where \textsc{\LayerVariantNameI} falls below 0.95* \textsc{Dense} are marked in {\color{red} Red}. \textsc{\LayerVariantNameI} falls below the 0.95\% mark w.r.t. \textsc{Dense} on only 3 zero-shot and 2 \textbf{GLUE+} tasks, falling above the mark  on all \textsc{GLUE+} aggregate tasks and \textsc{OpenLlm}}
        \label{tab:CrossBenchmarkPerfAllDyadVariantsPythia160Cross10Magisterial}
\end{table*}

\subsubsection{Complete Timing Results}

\begin{table}[!ht]
    \centering
    \begin{tabular}{|c|l|l|l|l|}
    \hline
        Model & Forward Pass & Backward Pass & Total & Total speedup ratio \\ \hline
        Dense & 96.57443477 & 218.1589193 & 315.6306277 & 1 \\ \hline
        \textsc{\LayerVariantNameI}-4 & 83.38802419 & 208.3585416 & 292.6851179 & 1.078396572 \\ \hline
        \textsc{\LayerVariantNameII}-4 & 82.48964827 & 207.7835725 & 291.2115524 & 1.083853388 \\ \hline
        \textsc{\LayerVariantNameIII}-4 & 83.34073742 & 210.0591608 & 294.3693217 & 1.072226636 \\ \hline
        \textsc{\LayerVariantNameI}-8 & 78.16424509 & 194.1724526 & 273.3341317 & 1.154742826 \\ \hline
    \end{tabular}{}
    \smallskip
    \caption{Mean time taken per minibatch by all modules of OPT-125m training on account of forward, backward passes and in total. All times are in milliseconds. Speedup ratio is computed w.r.t. \textsc{Dense}}
    \label{tab:OPT125m_AllModules_Time}
\end{table}

\begin{table}[!ht]
    \centering
    \begin{tabular}{|l|l|l|l|l|}
    \hline
        Model & Forward Pass & Backward Pass & Total & Total speedup ratio \\ \hline
        \textsc{Dense} & 2.548222502 & 4.971815463 & 7.520037964 & 1 \\ \hline
        \textsc{\LayerVariantNameI}-4 & 1.744403627 & 3.747922349 & 5.492325977 & 1.369190029 \\ \hline
        \textsc{\LayerVariantNameI}-8 & 1.111917225 & 3.026367151 & 4.138284376 & 1.817187337 \\ \hline
    \end{tabular}
    \smallskip
    \caption{Mean time taken per minibatch by the ff transformer modules of OPT-350m training on account of forward, backward passes and in total. All times are in milliseconds. Speedup ratio is computed w.r.t. 	\textsc{Dense}}
    \label{tab:OPT350m_FF_Time}
\end{table}

\subsubsection{Complete Memory Results}
In this section, we enclose complete results of evaluating each of our experiments along the aspects of memory \& parameter footprint we earmarked in \S\ref{subsec:benchmarksAndMetrics}


\begin{table}[!ht]
    \centering
    \begin{tabular}{|l|l|l|l|l|}
    \hline
        Model & Checkpoint Size (MB) & \# Params & In-Train GPU Use (MB) & \% Drop In GPU Mem vs Dense \\ \hline
        \textsc{Dense} & 478 & 86.63 & 9838 & 0 \\ \hline
        \textsc{Dyad-IT-4} & 370 & 58.32 & 9666 & 1.74832283 \\ \hline
        \textsc{Dyad-OT-4} & 370 & 58.32 & 9666 & 1.74832283 \\ \hline
        \textsc{Dyad-DT-4} & 370 & 58.32 & 9666 & 1.74832283 \\ \hline
        \textsc{Dyad-IT-8} & 316 & 44.16 & 9540 & 3.029070949 \\ \hline
    \end{tabular}
    \smallskip
    \caption{Mem./Param. Usage Metrics Across \textsc{Dense} and other {\LayerGroupName} variants for OPT-125m}
\end{table}

\subsubsection{Results on 100M Scale}

\begin{table*}[!t]
    \centering \scriptsize
    \begin{tabular}{| p{1.8cm} | p{1.2cm} | p{1.5cm} | p{1.5cm} | p{1.5cm}|} \hline
    Benchmark & Task & \textsc{Dense} & \textsc{Dense-Ext} & \textsc{\LayerVariantNameI} \\ \hline 
    \multirow{11}{*}{\textsc{GLUE+}}      & CoLA & 76.742 & 73.7 & 74.877 \\  
    & SST-2 & 87.992 & 86.6 & 89.567  \\ 
    & MPRC (F1) & 82.129 & 82.1 & 80.292  \\ 
    & QQP (F1) & 83.993 & 77.8 & 82.151 \\ 
    & MNLI & 77.339 & 70.1 & 76.623 \\  
    & MNLI-mm & 78.326 & 71.9 & 77.912 \\  
    & QNLI & 83.552 & 80.1 & 84.208 \\  
    & RTE & 53.535 & 67.7 & 63.366\\ 
    & BoolQ & 65.284 & 66.0 & 65.145\\  
    & MultiRC & 62.212 & 61.1 & 64.294        \\ 
    & WSC & 61.446 & 59.0 &  59.036 \\   \hline
    \multirow{3}{*}{Means}     &      \textsc{GLUE+} & 73.5808 & 72.24 & 74.2594 \\ 
    & \textsc{GLUE+-QA} & 69.875 & 69.7333 & 69.91033 \\ 
    & \textsc{GLUE+-NLI} & 73.188 & 72.45 & 75.52725 \\ \hline \hline 
    \multirow{17}{*}{\textsc{Blimp}}     & Anaphor Agr. & 97.90 & 94.9 & 90.03\\  
    & Agr. Structure & 77.885 & 73.8 & 78.48\\  
    & Binding & 72.306 & 73.8 & 73.89 \\  
    & Control/Raising & 74.105 & 72.2 & 72.67 \\  
    & D-N Agr. & 93.039 & 93.1 & 91.46 \\  
    & Ellipsis & 81.062  & 80.5 & 81.64 \\  
    & Filler-Gap & 74.214  & 73.6 & 74.167\\  
    & Irregular Forms & 89.924 & 80.8 & 85.55\\  
    & Island Effects & 62.780 & 57.8 & 57.81 \\  
    & NPI Licensing & 61.160  & 51.6 & 50.42\\  
    & Quantifiers & 71.303 & 74.5 & 67.46\\  
    & Subject Verb Agreement & 82.240 & 77.3 & 78.64 \\  
    & Hypernym & 47.791 & 46.3 & 47.56\\  
    & QA Congruence (Easy) & 70.312 & 76.5 & 76.56 \\  
    & QA Congruence (Tricky) & 52.121 & 47.9 & 50.91 \\  
    & Subject Auxiliary Inversion & 85.045 & 85.3 & 83.85\\  
    & Turn Taking & 79.643 & 82.9 & 79.28\\ \hline
    \multirow{1}{*}{Means}     & \textsc{Blimp} & 74.8723 & 73.1058 & 72.9633 \\
    \hline \hline
    \multirow{4}{*}{\textsc{OpenLlm}}     & ArcChallenge-25 & 25.256 & 23.293 & 24.659 \\
    & Hellaswag-10 & 25.234  & 25.055 & 25.473 \\
    & TruthfulQA-MC-0 &  48.868 & 48.448 & 49.332 \\
    & MMLU-5 & 23.567  & 23.181 & 23.080 \\ \hline
    \multirow{1}{*}{Means}     & \textsc{OpenLlm} & 30.73 & 29.99 & 30.636 \\ \hline
    \end{tabular}   
    \caption{Benchmark numbers for OPT-125m pretrained on \textsc{Str} (100M) comparing internal and external \textsc{Dense} baselines with Layer Variant}
\end{table*}

\subsection{Representational power of {\LayerGroupName}} \label{subsec:ReprPowerDyad}
Consider a network with two square {\LayerGroupName} layers i.e. ${{n}_{in}} = {{n}_{out}}$ sequentially applied one after the other to the input. Let the weight matrixes for the layers be, $W_1^{d}$ and $W_2^{d}$  and the input be $X$. The output $Y$ can be calculated as $Y = W_2^{d}W_1^{d}X$. 

Consider an input dimension $i$ of $X$ and an output dimension $j$ of $Y$. If $i//{{n}_{in}} = j//{{n}_{in}}$ i.e. they fall in the same block of the {\DyadComponentIShorthand} then there exists $O({{n}_{in}})$ connections between them through the middle layer. If  $i//{{n}_{in}} \ne j//{{n}_{in}}$ then only through the {\DyadComponentIShorthand} there wouldn't be any interactions between this pair of input and output. However, the {\DyadComponentIIShorthand} interacts with outputs that are spaced uniformly apart at a stride of ${{n}_{dyad}}$. On average $O({{n}_{in}}/{{n}_{dyad}})$ fall in the same block as that of the output dimension $j$, i.e. if the middle dimension was $k$ then $k//{{n}_{in}} = j//{{n}_{in}}$. Each of these middle dimensions will have a direct connection to j. Thus, in this second case the input dimension $i$ will have $O({{n}_{in}}/{{n}_{dyad}})$ connections to output dimension $j$. This is summarized in Eq \ref{DyadConnect}.

\begin{align}
    \text{No. of connections in Dyad} &= 
\begin{cases}
    O({{n}_{in}}),& \text{if } j//{{n}_{in}} = i//{{n}_{in}} \label{DyadConnect} \\
    O({{n}_{in}}/{{n}_{dyad}}),    & \text{otherwise}
\end{cases}
\end{align}

In the case of a sequence of two dense linear layers with the same shape, the number of connections would be $O({{n}_{in}} \times {{n}_{dyad}})$ between each input and output. Thus, the ratio of connections between dense and linear are as shown in Eq \ref{RatioConnect}.

\begin{align}
    \text{Ratio of connections in Linear to Dyad} &= 
\begin{cases}
    O({{n}_{dyad}}),& \text{if } j//{{n}_{in}} = i//{{n}_{in}} \label{RatioConnect} \\
    O({{n}_{dyad}}^2),    & \text{otherwise}
\end{cases}
\end{align}

Hence, {\LayerGroupName} layer has the ability to mix dimensions that are both near by and far away but the ability to mix information in nearby dimensions falls linearly with sparsity but far away dimensions fall quadratically. This means that {\LayerGroupName} will have a bias for pushing information that needs to interact with each other a lot close by and thus more efficiently using it's parameter space when compared to linear layers. Also the inter connections between the input and output dimensions fall gradually with {\DyadDim} and thus provides a way to tradeoff between representational power and computational cost.

\end{document}